\documentclass{article}

    \usepackage[preprint]{neurips_2026}

\usepackage[utf8]{inputenc} % allow utf-8 input
\usepackage[T1]{fontenc}    % use 8-bit T1 fonts
\usepackage{hyperref}       % hyperlinks
\usepackage{url}            % simple URL typesetting
\usepackage{booktabs}       % professional-quality tables
\usepackage{amsfonts}       % blackboard math symbols
\usepackage{nicefrac}       % compact symbols for 1/2, etc.
\usepackage{microtype}      % microtypography
\usepackage{xcolor}         % colors

\usepackage{graphicx}
\usepackage{subfigure}
\usepackage{pifont}
\usepackage{setspace} 
\usepackage{colortbl}
\usepackage{dsfont}
\usepackage{bbm}
\usepackage{multirow}
\usepackage{amsmath}
\usepackage{amssymb}
\usepackage{mathtools}
\usepackage{amsthm}
\RequirePackage{algorithm}
\RequirePackage{algorithmic}

\definecolor{light_green}{HTML}{E3F2D9}
\definecolor{deep_green}{HTML}{C8E5B3}
\definecolor{text_deep_green}{HTML}{72B73F}

\usepackage{hyperref}

\newcommand{\Checkmark}{\ding{51}}
\newcommand{\Crossmark}{\ding{55}}
\usepackage{pifont} % for \ding

\definecolor{mydarkblue}{rgb}{0,0.08,0.45}
\definecolor{darkred}{rgb}{0.55,0,0}

\hypersetup{
    colorlinks=true,
    citecolor=mydarkblue, % \cite 
    linkcolor=darkred    % \ref 
}

\title{
AttenA+: Rectifying Action Inequality in \\Robotic Foundation Models
}

\author{%
  Daojie Peng\textsuperscript{1}\thanks{Equal contribution.}\quad Fulong Ma\textsuperscript{1}\footnotemark[1]\quad Jiahang Cao\textsuperscript{2}\footnotemark[1]\quad Qiang Zhang\textsuperscript{1,3,6}\quad Xupeng Xie\textsuperscript{1} \\
    \textbf{Jian Guo\textsuperscript{4}\quad Ping Luo\textsuperscript{2}\quad Andrew F. Luo\textsuperscript{2}\quad Boyu Zhou\textsuperscript{5}\quad Jun Ma\textsuperscript{1}\thanks{Corresponding author: \texttt{jun.ma@ust.hk}}} \\
\textsuperscript{1}HKUST(GZ)\quad \textsuperscript{2}HKU\quad \textsuperscript{3}USTC \\
\textsuperscript{4}IDEA Research\quad \textsuperscript{5}SUSTech\quad \textsuperscript{6}X-Humaniod
}

\begin{document}

\maketitle

\begin{abstract}
\label{sec:abstract}

Existing robotic foundation models, while powerful, are predicated on an implicit assumption of \textit{temporal homogeneity}: treating all actions as equally informative during optimization. This "flat" training paradigm, inherited from language modeling, remains indifferent to the underlying physical hierarchy of manipulation. In reality, robot trajectories are fundamentally heterogeneous, where low-velocity segments often dictate task success through precision-demanding interactions, while high-velocity motions serve as error-tolerant transitions. Such a misalignment between uniform loss weighting and physical criticality fundamentally limits the performance of current Vision-Language-Action (VLA) models and World-Action Models (WAM) in complex, long-horizon tasks.
To rectify this, we introduce \textbf{AttenA+}, an architecture-agnostic framework that prioritizes kinematically critical segments via velocity-driven action attention. By reweighting the training objective based on the inverse velocity field, AttenA+ naturally aligns the model's learning capacity with the physical demands of manipulation. As a plug-and-play enhancement, AttenA+ can be integrated into existing backbones \textit{without structural modifications or additional parameters.}
Extensive experiments demonstrate that AttenA+ significantly elevates the ceilings of current state-of-the-art models. Specifically, it improves OpenVLA-OFT to 98.6\% (+1.5\%) on the \textsc{Libero} benchmark and pushes FastWAM to 92.4\% (+0.6\%) on RoboTwin 2.0. Real-world validation on a Franka manipulator further showcases its robustness and cross-task generalization. Our work suggests that mining the intrinsic structural priors of action sequences offers a highly efficient, physics-aware complement to standard scaling laws, paving a new path for general-purpose robotic control. 
The code is available at: \hyperlink{https://github.com/DaojiePENG/AttenA-Plus}{https://github.com/DaojiePENG/AttenA-Plus}.
\end{abstract}

\section{Introduction}
\label{sec:introduction}

Vision-Language-Action (VLA) and World-Action Models (WAM) have recently emerged as a powerful paradigm for end-to-end robotic control, enabling robots to interpret multimodal instructions and execute complex physical tasks \cite{kim2024openvla, black2024pi_0, peng2026structured}. 
However, this success masks a fundamental misalignment: while all linguistic tokens are assumed to be equally informative in standard NLP training, robotic actions are \textit{inherently heterogeneous} in their physical significance. 

In a typical manipulation trajectory, not all action steps are created equal. Consider the task of picking up a fragile object: the rapid motion of the arm toward the object is transitional and error-tolerant, whereas the final, slow-speed adjustment of the gripper is precision-demanding and task-critical. Currently, dominant training frameworks adopt a "flat" optimization objective, assigning \textit{identical learning weights} to every timestep regardless of its physical role \cite{brohan2022rt, zitkovich2023rt, team2024octo}. This uniform treatment forces models to waste representational capacity on trivial transitional segments, while under-optimizing the slow, high-precision actions that actually determine task success \cite{kim2024openvla, bu2025univla, black2024pi_0, cen2025worldvla, peng2025lovon, intelligence2025pi, cao2026compose}. Consequently, even the most advanced VLA models often struggle with last-centimeter precision in complex robotics tasks \cite{hejna2025robot,bi2025motus,li2026causal}.

To bridge this gap, we argue that the physical properties of an action, specifically its velocity, should dictate its importance during training. We propose \textbf{AttenA+}, a universal framework that introduces \textit{velocity-driven action attention} to reweight trajectory learning. Our key insight is simple yet effective: the end-effector's velocity serves as a natural inverse proxy for precision demand. By assigning different optimization priorities to different actions, AttenA+ aligns model training with the intrinsic physics of manipulation. As an architecture-agnostic enhancement, AttenA+ can be seamlessly plugged into any existing robotic backbone without structural modifications or additional parameters.

Our contributions are summarized as follows:
\textbf{1)} We identify and formalize the \textit{action inequality} inherent in robotic trajectories, exposing a fundamental bias in current foundation models where the uniform treatment of all actions leads to suboptimal optimization of physically critical steps.
\textbf{2)} We introduce \textbf{AttenA+}, a plug-and-play optimization framework that utilizes the inverse velocity field as a physical prior to reweight trajectory learning, effectively aligning the model's focus with kinematically demanding manipulation phases.
\textbf{3)} Extensive evaluations on \textsc{Libero} and RoboTwin benchmarks show that AttenA+ significantly elevates the performance ceilings of current state-of-the-art models. Furthermore, real-world experiments on a Franka manipulator demonstrate that our method provides superior robustness and success rates specifically during precision-critical motions where standard baselines frequently fail.

\begin{figure}
    \centering
    \includegraphics[width=1.0\linewidth]{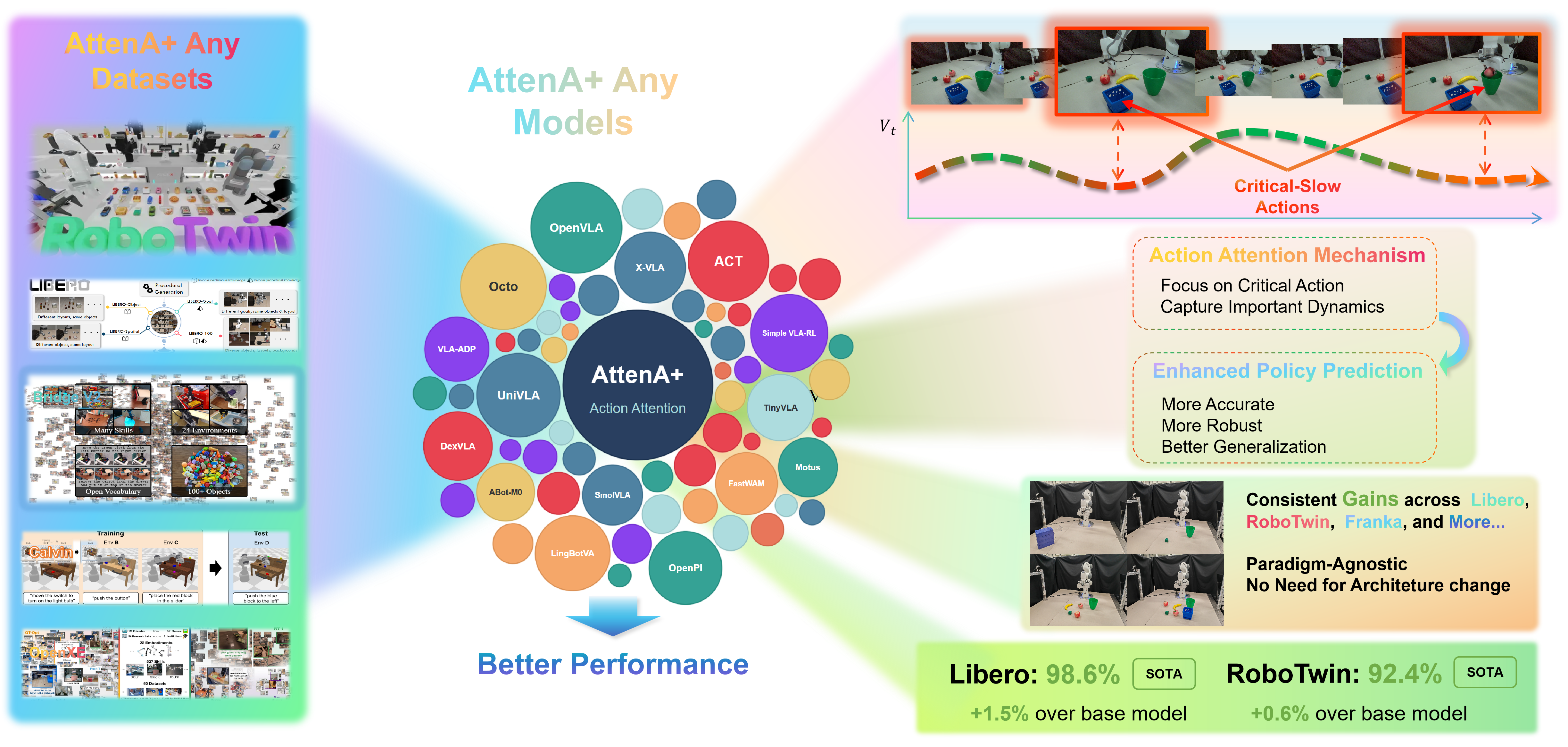}       
    \caption{\textbf{Overview of AttenA+}. AttenA+ is a paradigm-agnostic enhancement framework for action robotic foundation models, introducing velocity-field-based action attention to prioritize slow, critical manipulation steps. It seamlessly plugs into mainstream discriminative (e.g., OpenVLA-OFT) and generative ($\pi_0$, $\pi_{0.5}$, Diffusion Policy) architectures, as well as emerging World-Action Models (WAM). Without modifying core backbones or relying on data/model scaling, AttenA+ generalizes across diverse robotic datasets including \textsc{Libero} \cite{liu2023libero} and RoboTwin \cite{chen2025robotwin}, and consistently improves task success rates over state-of-the-art baselines.}
    \label{fig:basic_idea}
\end{figure}

\section{Related Works}
\label{sec:related_works}

\subsection{Robotic Foundation Models}
Vision-Language-Action (VLA) and World-Action Models (WAM) enable end-to-end robotic manipulation by grounding language in visual observations to generate continuous action sequences.
A wide range of VLA frameworks have been proposed to advance robotic manipulation performance, including foundational models and their variants, as well as specialized optimizations. OpenVLA \cite{kim2024openvla} serves as a core foundational framework unifying visual perception, language understanding, and action generation, with its variant OpenVLA-OFT \cite{kim2025fine} further optimizing via orthogonal fine-tuning to push state-of-the-art (SOTA) performance on \textsc{Libero} tasks. The $\pi$ model series, including $\pi_0$ \cite{black2024pi_0}, $\pi_0$ + FAST \cite{pertsch2025fast}, and $\pi_{0.5}$ \cite{intelligence2025pi}, advances generative VLA capabilities through flow matching for strong generalization. Other representative VLA models and optimizations include UniVLA \cite{bu2025univla}, VLA-ADP \cite{pei2025action}, CogACT \cite{li2024cogact}, SmolVLA \cite{shukor2025smolvla}, NORA and NORA-Long \cite{hung2025nora}, WorldVLA and WorldVLA* \cite{cen2025worldvla}, SP-VLA \cite{li2025sp}, FlashVLA \cite{tan2025think}, VLA-Cache \cite{xu2026vla}, FastV and FastV(+OFT) \cite{chen2024image}, SparseVLM \cite{zhang2024sparsevlm}, and CSP \cite{pei2024cross}.
% along with world modeling focused models such as Motus \cite{bi2025motus}, LingBot-VA \cite{li2026causal}, and Fast-WAM \cite{yuan2026fast}. 
Parallel efforts emerging as WAMs include Motus \cite{bi2025motus}, LingBot-VA \cite{li2026causal}, and Fast-WAM \cite{yuan2026fast}.
% These works collectively enhance VLA performance across diverse benchmarks but share a common limitation: treating all action timesteps equally during training, ignoring the inherent physical structure and varying importance of different motion phases.
Despite consistent progress across benchmarks, nearly all existing action models share a core limitation: treating all action timesteps equally during training,
neglecting the intrinsic physical hierarchy and heterogeneous importance of different motion phases.

\subsection{Action Sequence Modeling for Robotics}
Modeling sequential robotic actions is a core research direction, with early efforts focusing on trajectory optimization and inverse reinforcement learning (IRL). Recent data-driven approaches include Action Chunking with Transformers (ACT) \cite{zhao2023learning}, which uses transformers to model temporal dependencies in action sequences, and Diffusion Policy \cite{chi2025diffusion}, which leverages diffusion models for smooth, feasible trajectory generation—though these prioritize action quality over critical action prioritization based on physical characteristics (e.g., velocity).
Prior works have explored importance weighting for imitation learning: some weight entire trajectories by demonstration quality \cite{johns2021coarse,tangkaratt2020robust}, while others focus on per-timestep weighting (e.g., IRIS \cite{mandlekar2019iris} for informative offline data, uncertainty-based weighting for critical states \cite{wang2018exponentially}). However, these are limited to single-task learning or require extra overhead, unlike our velocity-based approach that needs no additional training and is compatible with arbitrary VLA frameworks. A small number of VLA works explore action weighting (e.g., VLA-ADP \cite{pei2025action} prunes redundant fast actions for efficiency), but none explicitly link velocity to learning priority. 
Different from all prior arts, \textbf{AttenA+} introduces a plug-and-play velocity-field weighting principle that emphasizes learning on critical action phases, requiring no extra supervision and universally compatible with mainstream robotic foundation models.

\subsection{Attention Mechanisms in Robotic Learning}
Attention has become a standard component in modern robotic foundation models, yet its usage remains largely confined to input modalities. Visual attention focuses on task-relevant spatial regions for manipulation \cite{huang2025spatial}; language attention aligns linguistic instructions with visual observations \cite{lynch2023interactive}; cross-modal attention further fuses vision and language features to predict actionable policies \cite{kim2024openvla, bu2025univla}. Remarkably, almost all prior attention designs operate exclusively on input vision-language streams, while output action trajectories are treated as plain unweighted regression targets.
Our work breaks this convention by proposing \textit{action attention}: we apply attention weighting directly to output action sequences, guided by physical velocity priors to emphasize precision-demanding motion segments. This extends attention design from input feature alignment to physical-aware action trajectory modeling, mirroring the hierarchical nature of human motor control.

% \subsection{Scaling Laws vs. Data Mining in Robotic Learning}The dominant paradigm for improving robotic models relies on scaling laws: expanding dataset size (e.g., RT-1 \cite{brohan2022rt} with 130k trajectories) or model capacity (e.g., PaLM-E \cite{driess2023palm} with 540B parameters). Inherited from NLP, this is inefficient for heterogeneous robotic action data, where critical and trivial actions coexist. Recent works challenge this paradigm (e.g., DemInf \cite{hejna2025robot} curates high-quality trajectories, physics-guided methods augment data). Our work advances this direction with a \textit{data mining} paradigm: instead of scaling, we mine velocity-based importance patterns in existing sequences to boost performance, offering a cost-effective alternative for data-constrained scenarios.

\section{Methodology}
\label{sec:method}

\subsection{The Homogeneity Bias in Robot Learning}
\label{sec:homogeneity_bias}

% Current robotic foundation models, regardless of their underlying architectures, typically treat the imitation of expert trajectories as a sequence of independent or autoregressive tokens with equal optimization priority. 
Current robotic foundation models, regardless of their underlying architectures, typically formulate expert trajectory as a sequence modeled via either independent single-step forecasting or autoregressive generation, with uniform optimization weight across all time steps.
This \textit{uniform weighting} strategy is prevalent across both discriminative and generative paradigms. Formally, given a dataset $\mathcal{D}$ of expert trajectories, the general optimization objective can be expressed as:
\begin{equation}
    \label{eq:general_loss}
    \theta^* = \arg\min_\theta \mathbb{E}_{\tau \sim \mathcal{D}} \left[ \sum_{t=1}^{T} \mathcal{L}_t (\pi_\theta(s_t), \boldsymbol{a}_t) \right],
\end{equation}
where $\mathcal{L}_t$ is the per-step loss function. In discriminative models \cite{kim2024openvla, bu2025univla}, $\mathcal{L}_t$ often takes the form of a regression loss (e.g., $L_1$ or $L_2$); in generative models such as diffusion policy \cite{chi2025diffusion,ze20243d,cao2025mamba} or flow matching models ($\pi_0$ \cite{black2024pi_0}), it corresponds to a score-matching or vector-field objective. 

Despite the diversity in loss formulations, these paradigms share an implicit assumption of \textit{temporal homogeneity}: every action token $a_t$ contributes identically to the overall gradient. However, this assumption is physically misaligned with the reality of robotic manipulation. By reducing complex physical interactions to a flat sequence of undifferentiated control signals, existing models inadvertently waste representational capacity on redundant transitional motions while under-optimizing the high-stakes, precision-demanding segments that truly govern task success.

\begin{figure*}[t]
    \centering
    \includegraphics[width=1.0\linewidth]{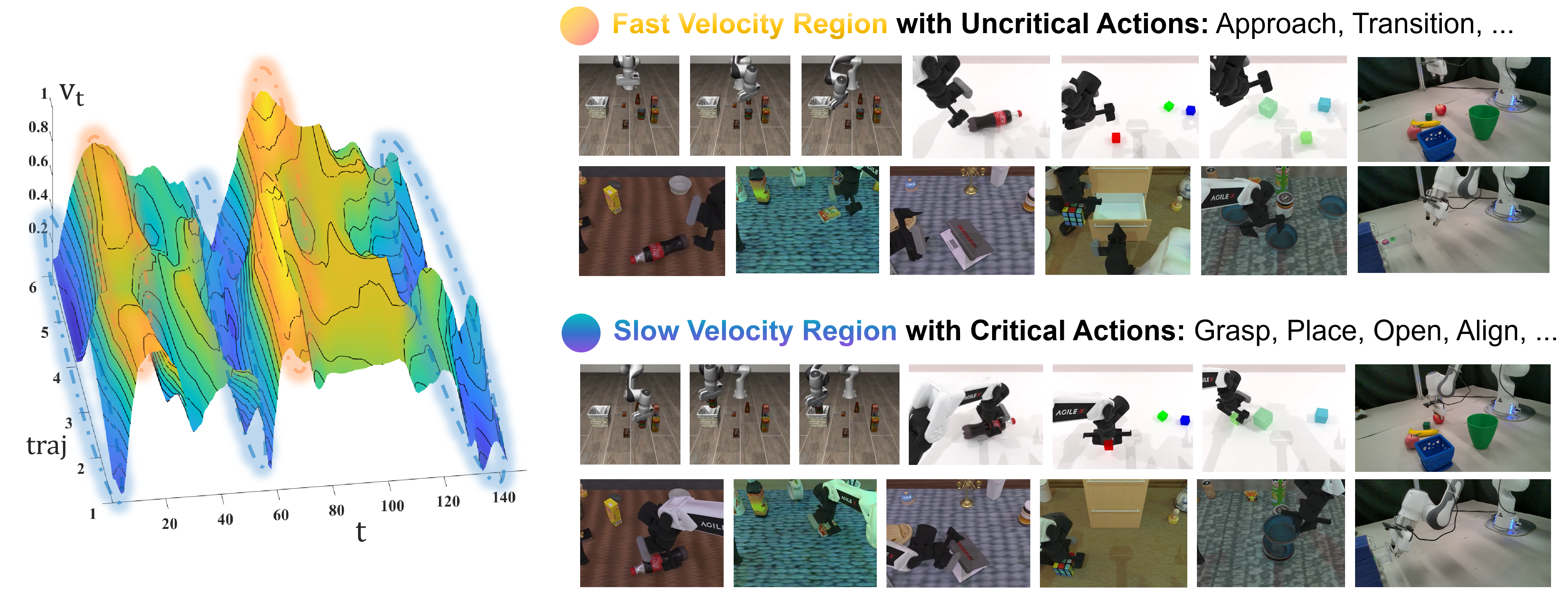}
    \caption{\textbf{Analysis of velocity fields revealing the inherent action inequality.} We observe that the informational density of the robot dataset is \textbf{non-uniformly distributed}: rapid motions are often redundant transitions, while slow-motion phases dominate task success or failure. The discovery of this kinematic hierarchy motivates the development of AttenA+, a plug-and-play mechanism designed to rectify the uniform weighting bias in current robotic foundation models.}
    \label{fig:velocity_field_vis}
\end{figure*}

\subsection{Quantifying Action Inequality via Velocity Fields}
\label{sec:action_inequality}

To rectify this misalignment, we propose a shift from uniform optimization toward \textit{Kinematic Criticality}. Our approach is rooted in the empirical discovery of \textit{Action Inequality}: the observation that the informational density of a manipulation sequence is non-uniformly distributed and is intrinsically linked to the movement velocity.

As visualized in Figure~\ref{fig:velocity_field_vis}, we analyze the velocity distribution across diverse task datasets. The results reveal a clear physical hierarchy within action sequences. High-velocity regions (highlighted in warm colors) typically correspond to "approach" or "transitional" phases—motions that occur in free space and are highly error-tolerant. In contrast, low-velocity regions (cold colors) consistently align with "interaction-rich" phases, such as precise alignment, grasping, or delicate placement. In these slow-motion segments, even a minor prediction error $\epsilon$ can lead to catastrophic task failure due to tight environmental constraints or contact dynamics.

We formalize this relationship by defining the instantaneous velocity magnitude $v_t$ of the ground-truth action $\boldsymbol{a}_t^{gt}$ at each timestep:
\begin{equation}
    \label{eq:velocity_def}
    v_t = \|\boldsymbol{a}_t^{gt}\|_2 = \sqrt{\sum_{d=1}^{D_{pos}} (a_{t,d}^{gt})^2},
\end{equation}
where $D_{pos}$ denotes the translational and/or rotational degrees of freedom. This metric $v_t$ serves as a natural, unsupervised proxy for task importance: lower velocity signifies higher precision demand. This discovery motivates a re-weighting of the optimization landscape to prioritize these low-velocity, high-criticality actions. In the following section, we introduce the \textbf{AttenA+} framework, which leverages this velocity-based prior to adaptively rescale the loss contribution of individual action tokens across different learning paradigms.

\begin{figure}
    \centering
    \includegraphics[width=1.0\linewidth]{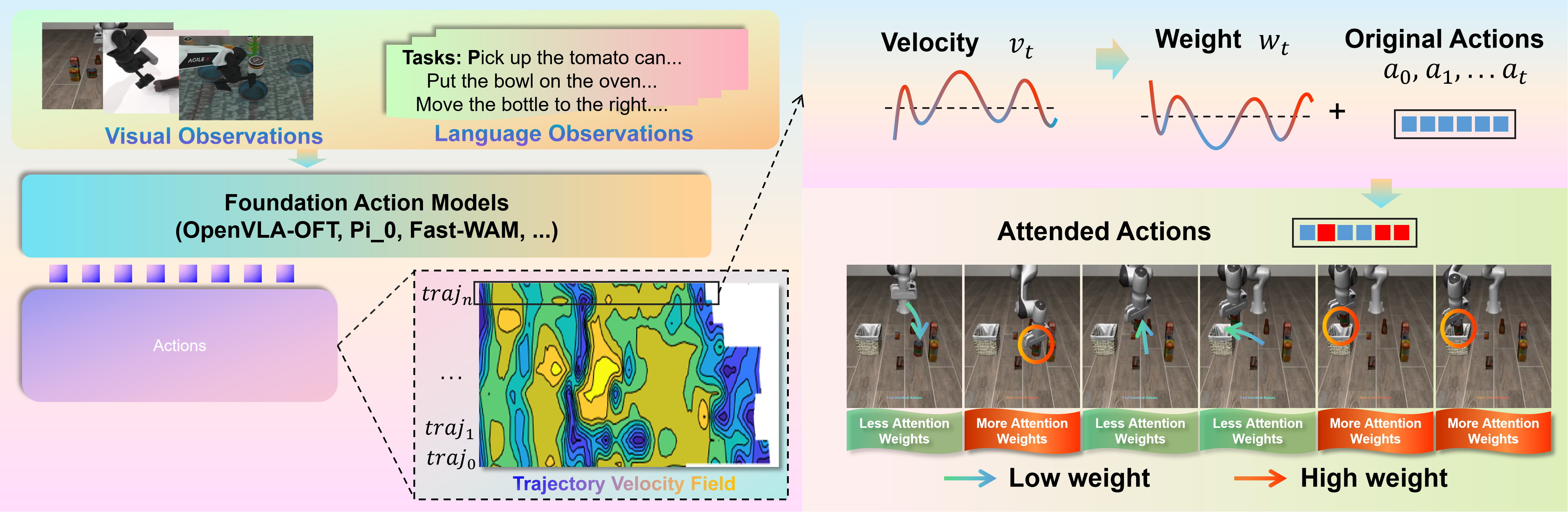}
    \caption{
    % Overview of AttenA+: Velocity-Field Attention for Robotic Action-Centric Models. (a) End-to-end workflow: given visual and language observations from datasets, we derive a velocity field. This field assigns higher attention weights to slow, critical manipulation steps and lower weights to fast transitional motions, prioritizing learning on error-sensitive actions while training the models. (b) Core mechanism: the velocity $v_t$ of each action $a_t$ is computed and mapped to an attention weight $w_t \propto 1/(v_t+\varepsilon)$, which is incorporated into the weighted loss $\mathcal{L}_{\text{AttenA+}}$. This ensures slow actions receive higher loss penalties, aligning model learning with the physical requirements of real-world manipulation.
    % AttenA+: Velocity-Field Attention for action-centric models. Workflow: a velocity field prioritizes slow, critical actions. Mechanism: velocity-aware weights are integrated into the weighted loss $\mathcal{L}_{\text{AttenA+}}$ to align training with manipulation physics.
    \textbf{Overview of AttenA+.} Given visual and language observations from datasets, we derive a velocity field. With attention weighting function $F_A$, this field assigns higher attention weights to slow, critical manipulation steps and lower weights to fast transitional motions, prioritizing learning on error-sensitive actions while training the models. 
    % This ensures slow actions receive higher loss penalties, aligning model learning with the physical requirements of real-world manipulation.
    }
    \label{fig:workflow}
\end{figure}

\subsection{Velocity-Field Attention (AttenA+)}
\label{sec:atten_a_plus}

To rectify the uniform weighting bias identified in current paradigms, we introduce \textbf{AttenA+}, a velocity-aware weighting mechanism designed to align model optimization with the physical criticality of robotic manipulation. As illustrated in Figure~\ref{fig:workflow}, AttenA+ functions as a "plug-and-play" enhancer that re-scales the loss manifold across diverse learning objectives without requiring architectural modifications.

\subsubsection{Weight Construction and Mapping}

The core of AttenA+ lies in translating the kinematic properties of expert demonstrations into an optimization priority. For a given dataset $\mathcal{D}$, we derive the instantaneous velocity magnitude $v_t$ following Equation~\ref{eq:velocity_def}. Taking the \textsc{Libero} benchmark ($D=7$) as a representative case, we compute the velocity magnitude using the first 6 dimensions (joint velocities) of the ground-truth action sequence $\mathcal{A}^{gt}$, omitting the binary gripper state to focus on continuous motion dynamics. This approach ensures that the resulting weight matrix $\mathcal{W} \in \mathbb{R}^{T \times 1}$ captures the intrinsic difficulty of the maneuver: low-speed segments, which consistently align with task-critical phases such as object grasping or precision placement, are assigned higher learning priorities, while high-speed transitional movements are downweighted.

We define the attention weighting function $F_A$ to map velocity to its corresponding importance weight:
\begin{equation}
    w_t = F_A(v_t).
\end{equation}
To accommodate varying task dynamics and noise profiles, we design four configurable mapping strategies: \textbf{inverse}, \textbf{inverse squared}, \textbf{exponential decay}, and \textbf{logarithmic}. These functions provide varying degrees of non-linear amplification for low-velocity actions, with detailed mathematical formulations provided in Appendix~\ref{appendix:weight}.

\subsubsection{Regularization for Training Stability}

Directly applying raw inverse velocity weights can lead to numerical instability or gradient dominance by near-static timesteps. To ensure robust convergence, AttenA+ incorporates two essential regularization steps:
\begin{itemize}
    \item \textbf{Weight Clipping}: We constrain the weights to a predefined range $[1/\text{clip}_{\text{max}}, \text{clip}_{\text{max}}]$. This prevents individual precision-critical steps from overwhelming the overall gradient and mitigates the impact of potential noise in expert demonstrations.
    \item \textbf{Loss Normalization}: We optionally normalize the weight vector such that $\frac{1}{T} \sum_{t=1}^T w_t \simeq 1$. This ensures that the global learning rate remains consistent with standard unweighted baselines, facilitating stable integration into existing training pipelines.
\end{itemize}

\subsubsection{Paradigm-Agnostic Optimization Objectives}

A defining advantage of AttenA+ is its \textbf{paradigm agnosticism}. It can be seamlessly integrated into diverse action models by augmenting the existing loss function.

\textbf{Discriminative Models (AttenA+Disc):} For standard regression-based VLAs, we transform the vanilla objective into a velocity-weighted $L_1$ loss:
\begin{equation}
    \label{eq:attenvla_discriminative}
    \theta^* = \arg\min_\theta \mathbb{E}_{(\mathcal{I}, L, \mathcal{A}^{gt}) \sim \mathcal{D}} \left[ \frac{1}{T \cdot D} \sum_{t=1}^T \sum_{d=1}^D w_t \cdot | a_{t,d}^{\text{pred}} - a_{t,d}^{gt} | \right],
\end{equation}
where $\theta$ denotes the model parameters and $w_t$ represents the velocity-derived weight.

\textbf{Flow Matching Models (AttenA+FM):} For generative frameworks such as $\pi_{0}$ or $\pi_{0.5}$, we revise the flow-matching objective to guide the model toward learning more accurate flow fields specifically for high-criticality segments:
\begin{equation}
    \label{eq:attenvla_fm_pi0_pi05}
    \phi^* = \arg\min_\phi \mathbb{E}_{\substack{(\mathcal{I}, L, \mathcal{A}^{gt}) \sim \mathcal{D} \\ \epsilon \sim \mathcal{N}(0, I)}} \left[ \frac{1}{T \cdot D} \sum_{t=1}^T \sum_{d=1}^D w_t \cdot \| u_t(\epsilon; \mathcal{I}, L) - (a_{t,d}^{gt} - \epsilon_d) \|_2^2 \right],
\end{equation}
where $u_t$ is the predicted flow field. By prioritizing these segments, AttenA+ enables generative models to capture the subtle nuances of precision-demanding actions that are often "washed out" in uniform training paradigms.

\section{Experiment}\label{sec:experiments}

We evaluate AttenA+ using four metrics:
(1) Success Rate (SR) (\%): percentage of successfully completed tasks.
(2) Average Success Rate ($\overline{\text{SR}}$) (\%): mean success rate across tasks.
(3) Average Error Rate ($\overline{\text{ER}}$) (\%): mean error rate across tasks.
(4) Average Success Rate Improvement ($\overline{\text{SR-I}}$) (\%): absolute gain in average success rate.
(5) Average Relative Error Rate Reduction ($\overline{\text{RER-R}}$) (\%): relative error reduction computed by
\begin{equation}
    \label{eq:rer_r}
    \overline{\text{RER-R}} = \Big(1 - \frac{\overline{\text{ER}}_{\text{AttenA+}}}{\overline{\text{ER}}_{\text{other}}}\Big) \times 100.
\end{equation}

\subsection{\textsc{Libero} and RoboTwin 2.0 Benchmark}
We build AttenA+OFT upon the official OpenVLA-OFT framework, and benchmark our approach on \textsc{Libero} dataset (Figure~\ref{fig:real_experiments}-I-(a)) against representative state-of-the-art VLA and WAM models across all four task subsets of the \textsc{Libero} dataset.
We select the best-performing checkpoint from training, then conduct evaluation across 4 random seeds to report the mean and standard deviation of success rates.
Additional training configurations are provided in Appendix~\ref{appendix:oft}.
As summarized in Table~\ref{tab:libero_results}, AttenA+OFT obtains an overall average success rate of 98.6\%, surpassing the prior SOTA OpenVLA-OFT by 1.5\%.
Consistent performance gains are observed across all task categories. In particular, our method achieves a 2.1\% improvement on long-horizon manipulation tasks, verifying that our action attention mechanism effectively enhances robustness and precision for complex, extended sequential behaviors.

\begin{table*}[]
    \centering
    \caption{Performance on \textsc{Libero} Compared with SOTA Methods.  $\overline{\text{SR}}$(\%): Average Success Rate; $\overline{\text{ER}}$(\%): Average Error Rate; $\overline{\text{SR-I}}$(\%): Average Success Rate Improvement; $\overline{\text{RER-R}}$(\%): Average Relative Error Rate Reduction (Compared with AttenA+OFT using Equation~\ref{eq:rer_r}).}
    \label{tab:libero_results}
    \resizebox{0.98\textwidth}{!}{
    \begin{tabular}{l|cccccccc}
    \toprule
        \textbf{Method} & \textbf{Spatial} & \textbf{Object} & \textbf{Goal} & \textbf{10} & \textbf{$\overline{\text{SR}}$ \textcolor{teal}{$\uparrow$}} & \textbf{$\overline{\text{ER}}$ \textcolor{red}{$\downarrow$}} & \textbf{$\overline{\text{SR-I}}$} & \textbf{$\overline{\text{RER-R}}$} \\
        \midrule
        OpenVLA \cite{kim2024openvla} & 84.7 & 88.4 & 79.2 & 53.7 & 76.50 & 23.50 & \textcolor{teal}{+22.1} & \textcolor{purple}{-94.0} \\
        SparseVLM \cite{zhang2024sparsevlm} & 79.8 & 67.0 & 72.6 & 39.4 & 64.70 & 35.30 & \textcolor{teal}{+33.9} & \textcolor{purple}{-96.0} \\
        FastV \cite{chen2024image} & 83.4 & 84.0 & 74.2 & 51.6 & 73.30 & 26.70 & \textcolor{teal}{+25.3} & \textcolor{purple}{-94.8} \\
        % FastV(+OFT) \cite{chen2024image} & 96.8 & 81 & 96.4 & 73 & 86.8 & 13.2 & \textcolor{teal}{+11.85} & \textcolor{purple}{-89.77} \\
        VLA-Cache \cite{xu2026vla} & 83.8 & 85.8 & 76.4 & 52.8 & 74.70 & 25.30 & \textcolor{teal}{+23.9} & \textcolor{purple}{-94.5} \\
        FlashVLA \cite{tan2025think} & 84.2 & 86.4 & 75.4 & 51.4 & 74.35 & 25.65 & \textcolor{teal}{+24.3} & \textcolor{purple}{-94.5} \\
        SP-VLA \cite{li2025sp} & 75.4 & 85.6 & 84.4 & 54.2 & 74.90 & 25.10 & \textcolor{teal}{+23.7} & \textcolor{purple}{-94.4} \\
        WorldVLA \cite{cen2025worldvla} & 85.6 & 89.0 & 82.6 & 59.0 & 79.05 & 20.95 & \textcolor{teal}{+19.6} & \textcolor{purple}{-93.3} \\
        % WorldVLA* \cite{cen2025worldvla} & 87.6 & 96.2 & 83.4 & 60 & 81.8 & 18.2 & \textcolor{teal}{+16.85} & \textcolor{purple}{-92.58} \\
        % NORA \cite{hung2025nora} & 85.6 & 87.8 & 77 & 45 & 73.85 & 26.15 & \textcolor{teal}{+24.8} & \textcolor{purple}{-94.84} \\
        NORA-Long \cite{hung2025nora} & 92.2 & 95.4 & 89.4 & 74.6 & 87.90 & 12.10 & \textcolor{teal}{+10.7} & \textcolor{purple}{-88.4} \\
        SmolVLA \cite{shukor2025smolvla} & 93.0 & 94.0 & 91.0 & 77.0 & 88.75 & 11.25 & \textcolor{teal}{+9.9} & \textcolor{purple}{-87.6} \\
        CogACT \cite{li2024cogact} & 97.2 & 98.0 & 90.2 & 88.8 & 93.55 & 6.45 & \textcolor{teal}{+5.1} & \textcolor{purple}{-78.3} \\
        CSP \cite{pei2024cross} & 84.7 & 82.2 & 77.1 & 74.3 & 79.58 & 20.42 & \textcolor{teal}{+19.1} & \textcolor{purple}{-93.1} \\
        $\pi_0$ + FAST \cite{pertsch2025fast} & 96.4 & 96.8 & 88.6 & 60.2 & 85.50 & 14.50 & \textcolor{teal}{+13.1} & \textcolor{purple}{-90.3} \\
        $\pi_0$ \cite{black2024pi_0} & 96.8 & 98.8 & 95.8 & 85.2 & 94.15 & 5.85 & \textcolor{teal}{+4.6} & \textcolor{purple}{-76.1} \\
        $\pi_{0.5}$ \cite{intelligence2025pi} & 98.8 & 98.2 & 98.0 & 92.4 & 96.85 & 3.15 & \textcolor{teal}{+1.8} & \textcolor{purple}{-55.6} \\
        UniVLA \cite{bu2025univla} & 96.5 & 96.8 & 95.6 & 92.0 & 95.23 & 4.77 & \textcolor{teal}{+3.4} & \textcolor{purple}{-70.7} \\
        VLA-ADP \cite{pei2025action} & \textbf{99.0} & 98.2 & 96.8 & 91.2 & 96.30 & 3.70 & \textcolor{teal}{+2.3} & \textcolor{purple}{-62.2} \\
        OpenVLA-OFT \cite{kim2025fine} & 97.6 & 98.4 & 97.9 & 94.5 & 97.10 & 2.90 & \textcolor{teal}{+1.5} & \textcolor{purple}{-51.7} \\
        \midrule
        % \textbf{AttenA+$\mathbf{\pi_{0.5}}$ (ours)}& \textbf{99.2 $\pm$ 0.18} & 99.6 $\pm$ 0.20 & 98.8 $\pm$ 0.26 & 94.2 $\pm$ 0.28 & 97.95 & 2.05 & - & - \\
        \textbf{AttenA+OFT (ours)} & \textbf{99.0 $\pm$ 0.16} & \textbf{100 $\pm$ 0.00} & \textbf{98.8 $\pm$ 0.28} & \textbf{96.6 $\pm$ 0.30} & \textbf{98.60} & \textbf{1.40} & - & - \\
        \bottomrule
    \end{tabular}}
\end{table*}

% \subsection{RoboTwin Benchmark}
We further validate our method on the RoboTwin benchmark (Figure~\ref{fig:real_experiments}-I-(b)), implementing AttenA+WAM based on the Fast-WAM framework. As shown in Table \ref{tab:robotwin}, AttenA+WAM achieves a new state-of-the-art average success rate of 92.46\%, improving the base model Fast-WAM by 0.6\% and outperforming the prior best LingBot-VA by 0.3\%, without requiring any embodied pre-training. This confirms that our action attention mechanism can effectively boost performance even on larger, more diverse real-world benchmarks.

\begin{table}[t]
\centering
\caption{Performance on RoboTwin 2.0 Compared with SOTA Methods. 
% Refer to Appendix~\ref{appendix:robotwin} for detailed results on all 50 tasks.
}
\resizebox{0.8\textwidth}{!}{
\begin{tabular}{l|c|cccccc}
    \toprule
    Method & Embodied PT. & Clean & Rand. & \textbf{$\overline{\text{SR}}$ \textcolor{teal}{$\uparrow$}} &  \textbf{$\overline{\text{ER}}$ \textcolor{red}{$\downarrow$}} & \textbf{$\overline{\text{SR-I}}$} & \textbf{$\overline{\text{RER-R}}$}  \\
    \midrule
    $\pi_0$ \cite{black2024pi_0} & \Checkmark & 65.92 & 58.40 & 62.20 & 37.80 & \textcolor{teal}{+30.3} & \textcolor{purple}{-80.1} \\
    $\pi_{0.5}$ \cite{intelligence2025pi} & \Checkmark & 82.74 & 76.76 & 79.75 & 20.25 & \textcolor{teal}{+12.7} & \textcolor{purple}{-62.8} \\
    X-VLA \cite{zheng2025x} & \Checkmark & 72.90 & 72.80 & 72.85 & 27.15 & \textcolor{teal}{+19.6 }& \textcolor{purple}{-72.2} \\
    Motus \cite{bi2025motus} & \Checkmark & 88.66 & 87.02 & 87.80 & 12.20 & \textcolor{teal}{+4.6 }& \textcolor{purple}{-38.0} \\
    LingBot-VA \cite{li2026causal} & \Checkmark & \underline{92.90} & 91.50 & \underline{92.2} & \underline{7.80} & \textcolor{teal}{+0.3 }& \textcolor{purple}{-3.3}\\
    Fast-WAM \cite{yuan2026fast} & \Crossmark & 91.88 & \underline{91.78} & 91.80 & 8.20 & \textcolor{teal}{+0.6 }& \textcolor{purple}{-7.7}\\
    \midrule
    \textbf{AttenA+WAM (ours)} & \Crossmark & \textbf{93.06} & \textbf{91.86} & \textbf{92.46} & \textbf{7.54} & - & -\\
    \bottomrule
\end{tabular}}
\label{tab:robotwin}
\end{table}

\subsection{Improvement of Different Models with Action Attention}
\label{subsec:improvement_diff_models}

\begin{figure}
    \centering
    \includegraphics[width=1.0\linewidth]{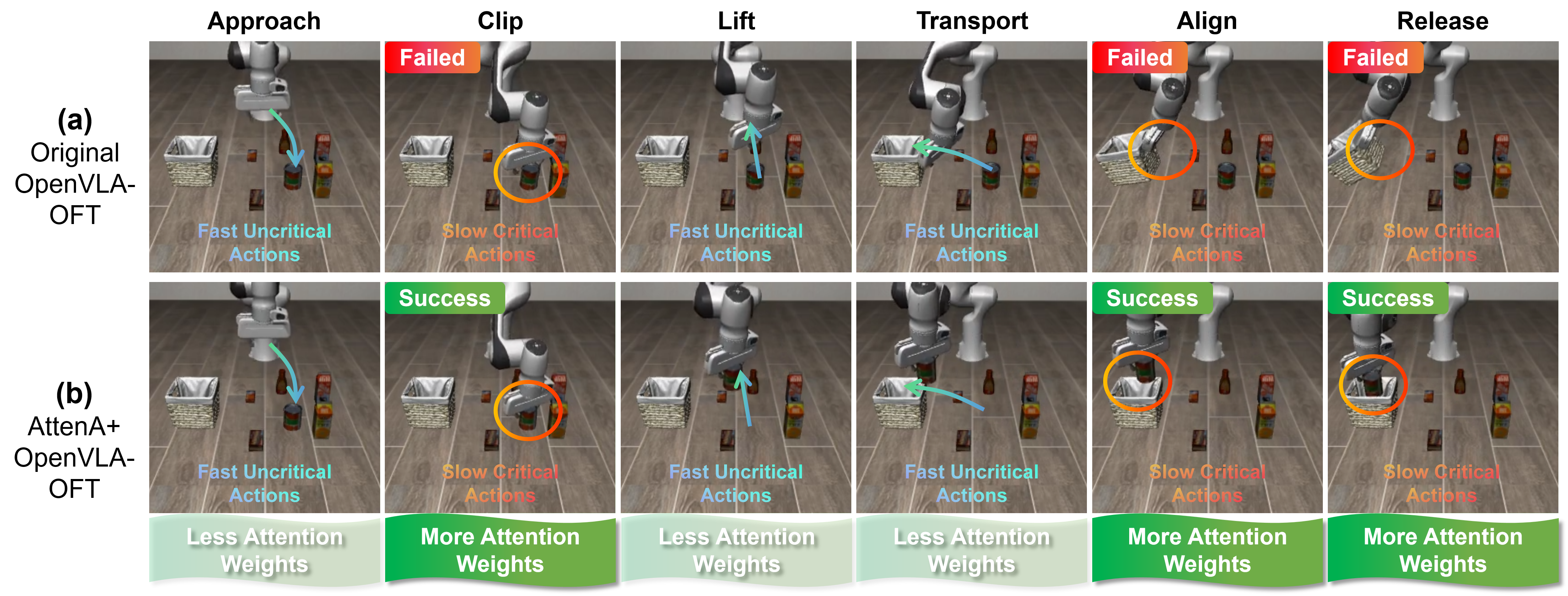}
    \caption{\textbf{Qualitative comparison of task execution with/without AttenA+.} (a) The original baseline fails due to accumulated errors in slow, critical manipulation steps (clip, align, release), which receive equal loss weight to fast transitional motions. (b) AttenA+ prioritizes these high-precision segments with larger attention weights, leading to stable task completion.}
    \label{fig:compare_demo}
\end{figure}

As shown in Table \ref{tab:model_perf_improve}, we validate the effectiveness and generality of our velocity-field-based action attention by integrating it into both discriminative and generative models. Figure \ref{fig:compare_demo} provides a qualitative comparison: the original baseline fails due to accumulated errors in slow, critical manipulation steps (clip, align, release), where precision is essential but receives equal loss weight to fast transitional motions. In contrast, AttenA+ prioritizes these high-precision segments with larger attention weights, leading to stable task completion.

For the discriminative framework, we apply our method to OpenVLA-OFT, a strong baseline already achieving high performance on the \textsc{Libero} benchmark. Equipped with action attention, AttenA+OFT yields consistent gains across all task categories: Spatial (+1.4\%), Object (+1.6\%), Goal (+0.9\%), and Long-horizon tasks (+2.1\%). The overall average success rate improves by +1.5\% (from 97.1\% to 98.6\%), with a corresponding -1.5\% reduction in error rate.
For the generative framework, we adopt $\pi_{0.5}$ as the backbone and construct AttenA+$\pi_{0.5}$. Similarly, consistent improvements are observed across all task types, with an average success rate increase of +1.10\%.

These results demonstrate that our velocity-field action attention is \textit{paradigm-agnostic} and can serve as a universal plug-and-play enhancement for both discriminative and generative models. Notably, the performance gain is most pronounced on long-horizon tasks, where distinguishing critical actions from transitional movements is essential for maintaining execution success.

% \begin{table*}[htbp]
%   \centering
%   \caption{Model Performance Improvement with Velocity-Field-Based Action Attention on LIBERO dataset.}
%   \label{tab:model_perf_improve}
%   \resizebox{0.90\textwidth}{!}{
%   \begin{tabular}{lc|cccccc}
%   \toprule
%   \multicolumn{2}{c}{\textbf{Model}} & \textbf{Spatial} & \textbf{Object} & \textbf{Goal} & \textbf{10} & $\overline{\textbf{SR}}$ & $\overline{\textbf{ER}}$ \\
%   \midrule
%   \multirow{3}{*}{\textbf{Generative}} 
%     & $\pi_{0.5}$    & 98.8 & 98.2 & 98 & 92.4 & 96.85 & 3.15 \\
%     & AttenA+$\pi_{0.5}$ & 99.2   & 99.6   & 98.8   & 94.2   & 97.95  & 2.05  \\
%     % \hline
%     & Improvement  & \textcolor{teal}{+0.4 } & \textcolor{teal}{+1.4 } & \textcolor{teal}{+0.8 } & \textcolor{teal}{+1.8 } & \textcolor{teal}{+1.10}  & \textcolor{purple}{-1.10} \\
%   \midrule
%   \multirow{3}{*}{\textbf{Discriminative}} 
%     & OpenVLA-OFT  & 97.6 & 98.4 & 97.9 & 94.5 & 97.1  & 2.9  \\
%     & AttenA+OFT    & 99.0   & 100 & 98.8   & 96.6 & 98.6 & 1.4 \\
%     % \hline
%     & Improvement  & \textcolor{teal}{+1.4 } & \textcolor{teal}{+1.6 } & \textcolor{teal}{+0.9 } & \textcolor{teal}{+2.1 } & \textcolor{teal}{+1.5}  & \textcolor{purple}{-1.5} \\
%   \bottomrule
%   \end{tabular}
%   }
% \end{table*}

\begin{table*}[htbp]
  \centering
  \caption{Performance Improvement with Velocity-Field-Based Action Attention on \textsc{Libero} dataset.}
  \label{tab:model_perf_improve}
  \resizebox{0.98\textwidth}{!}{
  \begin{tabular}{lc|cccccc}
  \toprule
  \multicolumn{2}{c}{\textbf{Model}} & \textbf{Spatial} & \textbf{Object} & \textbf{Goal} & \textbf{10} & $\overline{\textbf{SR}}$ & $\overline{\textbf{ER}}$ \\
  \midrule
  \multirow{2}{*}{\textbf{Generative}} 
    & $\pi_{0.5}$    & 98.8 & 98.2 & 98.0 & 92.4 & 96.85 & 3.15 \\
    & \textbf{AttenA+$\mathbf{\pi_{0.5}}$} & 99.2 (\textcolor{teal}{+0.4})   & 99.6 (\textcolor{teal}{+1.4})   & 98.8 (\textcolor{teal}{+0.8})   & 94.2 (\textcolor{teal}{+1.8})   & 97.95 (\textcolor{teal}{+1.10})  & 2.05 (\textcolor{purple}{-1.10})  \\
  \midrule
  \multirow{2}{*}{\textbf{Discriminative}} 
    & OpenVLA-OFT  & 97.6 & 98.4 & 97.9 & 94.5 & 97.1  & 2.9  \\
    & \textbf{AttenA+OFT}    & 99.0 (\textcolor{teal}{+1.4})   & 100 (\textcolor{teal}{+1.6}) & 98.8 (\textcolor{teal}{+0.9})   & 96.6 (\textcolor{teal}{+2.1}) & 98.6 (\textcolor{teal}{+1.50})  & 1.4 (\textcolor{purple}{-1.50}) \\
  \bottomrule
  \end{tabular}
  }
\end{table*}

\subsection{Ablation Study on Weighting Strategies and Clipping Thresholds}
\label{subsec:ablation_weight}

We conduct an ablation study to validate the effectiveness of our proposed velocity-based weighting strategies and the criticality of the weight clipping threshold $\text{clip}_{\text{max}}$, with OpenVLA-OFT as our baseline model.
Results on the \textsc{Libero} benchmark are reported in Table~\ref{tab:ablation_weight_strategy}.

First, we observe that \textit{no single weighting strategy universally dominates all task categories}, which aligns with the distinct motion characteristics of different robotic manipulation tasks.
Specifically, \texttt{inverse\_squared} achieves the best performance on \textsc{Libero-Spatial}, \texttt{inverse} performs optimally on \textsc{Libero-Object} and $\text{clip}_{\text{max}}$=$3.0$ settings of \textsc{Libero-Goal}, while \texttt{exp\_decay} and \texttt{log} show strong advantages on \textsc{Libero-10} and $\text{clip}_{\text{max}}$=$2.0$ settings of \textsc{Libero-Goal}.
This demonstrates that different velocity-aware weighting functions adapt to task-specific motion patterns.

Second, the clipping threshold $\text{clip}_{\text{max}}$ plays a vital role in balancing weight emphasis and training stability.
When $\text{clip}_{\text{max}}$=$1.0$, all weighted loss terms degenerate to the uniform baseline, yielding identical performance to the original OpenVLA-OFT.
As $\text{clip}_{\text{max}}$ increases to $2.0$ or $3.0$, our AttenA+ mechanism consistently improves the task success rates.
However, an overlarge threshold ($\text{clip}_{\text{max}}$=$5.0$) tends to degrade performance, as extreme weights introduce training instability and over-emphasize noisy low-velocity actions.
These results confirm that appropriate weight clipping is essential for maintaining the effectiveness of our velocity-field attention mechanism.

\begin{table}[t]
\centering
\caption{Ablation study on different velocity weighting strategies and weight clipping thresholds $clip_{\text{max}}$. We report task success rate (\%) on the \textsc{Libero} benchmarks. Baseline is OpenVLA-OFT.}
\label{tab:ablation_weight_strategy}
\resizebox{\linewidth}{!}{
\begin{tabular}{lcccccccccccc}
\toprule
& \multicolumn{2}{c|}{Libero-Spatial}
& \multicolumn{2}{c|}{Libero-Object}
& \multicolumn{2}{c|}{Libero-10}
& \multicolumn{6}{c}{Libero-Goal} \\
\textbf{Strategy / Clip$_{\text{max}}$}
& \multicolumn{2}{c|}{2.0}
& \multicolumn{2}{c|}{2.0}
& \multicolumn{2}{c|}{2.0}
& \multicolumn{2}{c|}{2.0}
& \multicolumn{2}{c|}{3.0}
& \multicolumn{2}{c}{5.0} \\
& SR & $\Delta$SR & SR & $\Delta$SR & SR & $\Delta$SR & SR & $\Delta$SR & SR & $\Delta$SR & SR & $\Delta$SR \\
\midrule
\textit{baseline}
& \textit{97.6} & -
& \textit{98.4} & -
& \textit{94.5} & -
& \textit{97.9} & -
& \textit{97.9} & -
& \textit{\textbf{97.9}} & - \\ \midrule
exp\_decay ($w_{b,t} = e^{-\alpha \cdot v_{b,t}}$)
& 99.2 & \textcolor{teal}{+1.6}
& 99.8 & \textcolor{teal}{+1.4}
& \textbf{96.8} & \textcolor{teal}{+2.3}
& \textbf{99.0} & \textcolor{teal}{+1.1}
& 95.4 & \textcolor{purple}{-2.5}
& 97.4 & \textcolor{purple}{-0.5} \\
inverse\_squared ($w_{b,t} = \frac{1}{v_{b,t}^2}$)
& \textbf{99.4} & \textcolor{teal}{+1.8}
& 99.8 & \textcolor{teal}{+1.4}
& 94.2 & \textcolor{purple}{-0.3}
& 98.8 & \textcolor{teal}{+0.9}
& 97.9 & 0.0
& 97.6 & \textcolor{purple}{-0.3} \\
inverse ($w_{b,t} = \frac{1}{v_{b,t}}$)
& 98.6 & \textcolor{teal}{+1.0}
& \textbf{100.0} & \textcolor{teal}{+1.6}
& 95.8 & \textcolor{teal}{+1.3}
& 98.0 & \textcolor{teal}{+0.1}
& \textbf{98.2} & \textcolor{teal}{+0.3}
& 95.6 & \textcolor{purple}{-2.3} \\
log ($w_{b,t} = \frac{1}{\log(1 + v_{b,t})}$)
& 98.2 & \textcolor{teal}{+0.6}
& 99.6 & \textcolor{teal}{+1.2}
& 88.8 & \textcolor{purple}{-5.7}
& \textbf{99.0} & \textcolor{teal}{+1.1}
& 97.8 & \textcolor{purple}{-0.1}
& 97.8 & \textcolor{purple}{-0.1} \\
\bottomrule
\end{tabular}
}
\end{table}

% \begin{table}[t]
% \centering
% \caption{Ablation study on different velocity weighting strategies and weight clipping thresholds $clip_{\text{max}}$. We report task success rate (\%) on the Libero benchmarks.}
% \label{tab:ablation_weight_strategy}
% \resizebox{\linewidth}{!}{
% \begin{tabular}{lcccccc}
% \toprule
% & \textbf{Libero-Spatial} & \textbf{Libero-Object} & \textbf{Libero-10} & \textbf{Libero-Goal} & \multicolumn{2}{c}{\textbf{$clip_{\text{max}}$}} \\
% \cmidrule(l){6-7}
% $F_\text{A}$ / $clip_{\text{max}}$ & 2.0 & 2.0 & 2.0 & 2.0 & 3.0 & 5.0 \\
% \midrule
% \textit{baseline}          & \textit{97.6} & \textit{98.4} & \textit{94.5} & \textit{97.9} & \textit{97.9} & \textit{97.9} \\
% \hline
% exp\_decay        & 99.0 & 99.8 & 96.8 & 99.0 & 95.4 & 97.4 \\
% inverse\_squared  & 99.4 & 99.8 & 51.2 & 98.8 & 97.9 & 97.6 \\
% inverse           & 98.6 & 100  & 95.8 & 98.0 & 98.2 & 95.6 \\
% log               & 98.2 & 99.6 & 58.8 & 99.0 & 97.8 & 97.8 \\
% \bottomrule
% \end{tabular}
% }
% \end{table}

\subsection{Real-World Robot Experiments}
\begin{figure}
    \centering
    \includegraphics[width=1.0\linewidth]{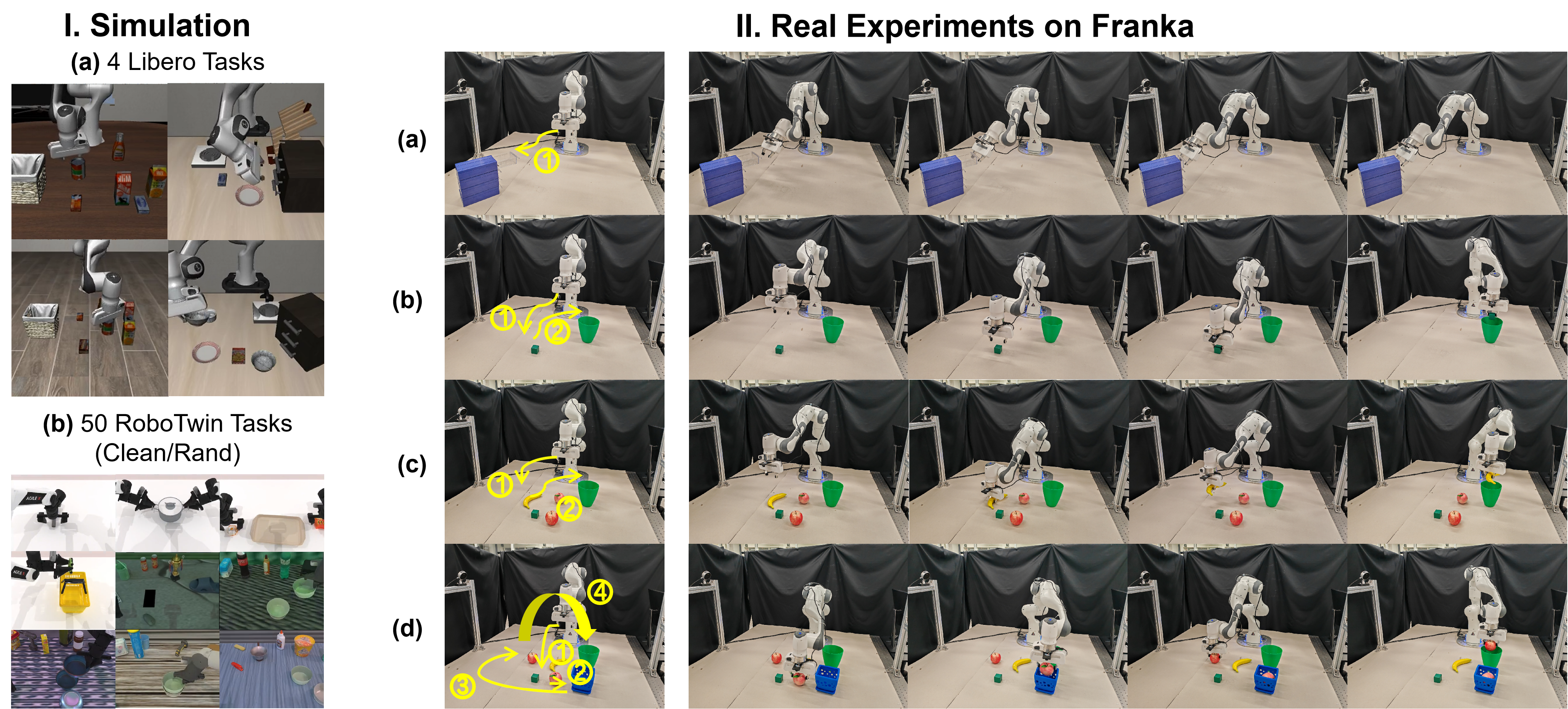}
    \vspace{-3mm}
    \caption{\textbf{Overview of experimental tasks.} I. Simulation: (a) Four \textsc{Libero} benchmark tasks; (b) 50 diverse RoboTwin tasks, including clean and randomized environments. II. Real-world experiments on Franka Panda: (a)--(d) Four representative tasks (drawer opening, pick-and-place, multi-objects, and sequential manipulation), showing AttenA+ enhanced policy execution.}
    \label{fig:real_experiments}
\end{figure}

As shown in Figure \ref{fig:real_experiments}, we design 4 kinds of task for validation using the Franka manipulator: \textit{(a) Close the open drawer, (b) Put the Green Cube into Green Bowl, (c) Put Object-A into Green Bowl, (d) Put Object-A into XXX and then put Object-B into XXX.} For the easy tasks (a) and (b), we collect 50 trajectories for demonstration. For harder tasks (c) and (d), we collect 100 trajectories for demonstration. Notably, during demonstration data collection, we use different speed for different phase: at the beginning, we use the baseline speed for approaching the object for grasping, then we change the speed to be 1/3 of the baseline to fine align and operate the object which indicates critical actions. Then after grasping the object we change the speed to baseline and fastly move to the bowl. When approaching the bowl, the speed is again reduced to 1/3 of the baseline for fine align to the bowl and finally release the object. After collection, we clean the trajectory by removing the no action waiting frames and do action smoothing for efficient training and action attention. 

\begin{figure}[t]
  \centering
  \begin{minipage}{0.48\linewidth}
    \centering
    % \caption{table}{Real robot experiments on the Franka platform. Each task is tested over 50 trials.}
    % 缩放表格，防止被覆盖
    \resizebox{\linewidth}{!}{
    \begin{tabular}{c|cccc}
      \toprule
      Model & Close Draw & Put Cube & Multi-object & Long \\
      \midrule
      OpenVLA-OFT  & 50/50 & 48/50 & 45/50 & 42/50 \\
      AttenA+OFT   & 50/50 & 50/50 & 49/50 & 45/50 \\
      Improvement  & \textcolor{teal}{0\%} & \textcolor{teal}{4$\% \uparrow$} & \textcolor{teal}{8$\% \uparrow$} & \textcolor{teal}{6$\% \uparrow$} \\
      \bottomrule
    \end{tabular}
    }
    \label{tab:real_experiments_results}
  \end{minipage}
  \hfill
  \begin{minipage}{0.48\linewidth}
    \centering
    \includegraphics[width=\linewidth]{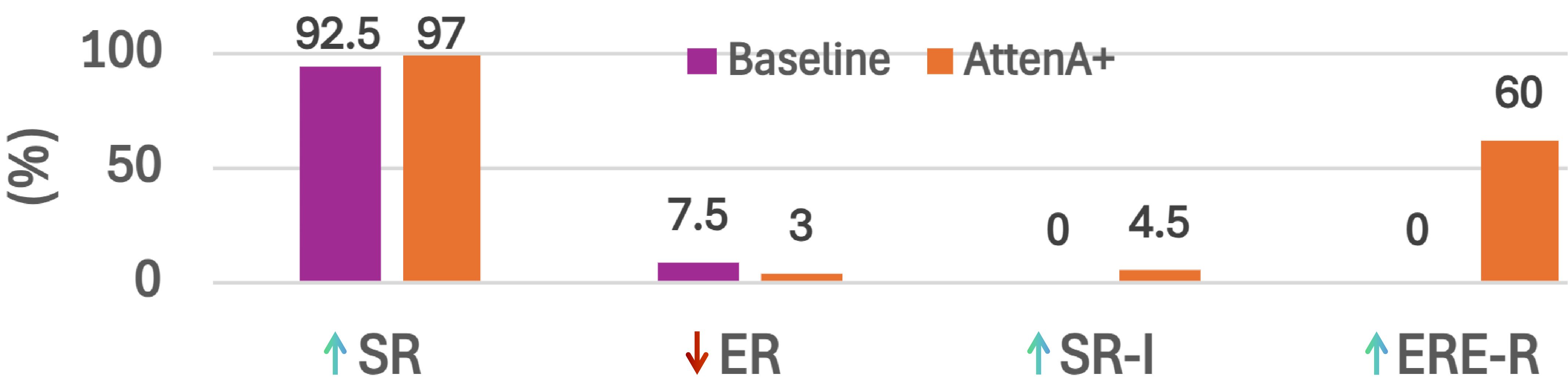}
  \end{minipage}
  \caption{\textbf{Real robot experimental results on Franka} (Each task is tested over 50 trials): (a) Quantitative success rates (\%); (b) Qualitative performance visualization.}
  \label{fig:real_overall}
\end{figure}

We then finetune and test the task following the OpenVLA-OFT recipe with 2 Nvidia H800 GPUs.
In the testing phase, we deploy the model on a RTX-4090 GPU, evaluate each task for 50 times and compute the $\overline{\textbf{SR}}$. The results are shown in Figure~\ref{fig:real_overall}.
% Table \ref{tab:real_experiments_results}. 
We can see that AttenA+OFT consistently outperforms the baseline OpenVLA-OFT across all real-world tasks, improving the average success rate from 92.5\% to 97.0\%, with the largest gains on the more complex multi-object and long-horizon tasks, further validating the effectiveness of our method in real-world scenarios.

% \begin{table}[]
%     \centering
%     \caption{Real Experiments on Franka. Each test is repeated 50 times and compute the success rate.}
%     \begin{tabular}{c|cccccc}
%         \toprule
%         Model & Close Draw & Put Cube & Multi-object & Long & $\overline{\textbf{SR}}$ & $\overline{\textbf{ER}}$\\
%         \midrule
%         OpenVLA-OFT  & 100 & 96 & 90 & 84 & 92.5 & 7.5\\
%         AttenA+OFT  & 100 & 100 & 98 & 90 & 97.0 & 3.0\\
%         Improvement  & \textcolor{teal}{0} & \textcolor{teal}{4} & \textcolor{teal}{8} & \textcolor{teal}{6} & \textcolor{teal}{4.5} & \textcolor{purple}{-4.5}\\
%         \bottomrule
%     \end{tabular}
    
%     \label{tab:real_experiments_results}
% \end{table}

% \begin{table}[]
%     \centering
%     \caption{Real Experiments on Franka. Each test is repeated 50 times and compute the success rate.}
%     \begin{tabular}{c|cccc}
%         \toprule
%         Model & Close Draw & Put Cube & Multi-object & Long \\
%         \midrule
%         OpenVLA-OFT  & 100 & 96 & 90 & 84 \\
%         AttenA+OFT  & 100 & 100 & 98 & 90 \\
%         Improvement  & \textcolor{teal}{0} & \textcolor{teal}{4} & \textcolor{teal}{8} & \textcolor{teal}{6} \\
%         \bottomrule
%     \end{tabular}
    
%     \label{tab:real_experiments_results}
% \end{table}

% \begin{figure}
%     \centering
%     \includegraphics[width=1.0\linewidth]{figures/real_bins.png}
%     \caption{Real Experiments Results on Franka.}
%     \label{fig:real_bins}
% \end{figure}

\section{Conclusion}
\label{sec:conclusion}

This work presents \textbf{AttenA+}, a generic enhancement framework for robotic foundation models. It introduces velocity-field-based action attention to prioritize critical, low-speed manipulation steps during training, aligning model optimization with real-world manipulation physics without modifying core architectures. Evaluated on \textsc{Libero} and RoboTwin 2.0 benchmarks, AttenA+ consistently improves success rates and reduces errors across both discriminative and generative paradigms (including VLA and WAM), and is readily extendable to other architectures such as diffusion policies.

We also note two main limitations. First, our velocity-weighted attention relies on a hand-crafted heuristic, which assumes critical manipulation steps are inherently slow. This does not generalize to dynamic tasks (e.g., high-speed grasping, table tennis) where critical actions may instead be fast and ballistic. Second, the mechanism only leverages velocity information, ignoring other physical cues such as force or torque that can signal action importance.

Future work will move beyond fixed heuristics toward \emph{physically grounded, learnable action attention} that integrates multi-modal physical signals and adapts dynamically to task semantics. By respecting the structure of robotic actions rather than treating all timesteps equally, we can build more efficient, robust, and generalizable robotic foundation systems.

{\small
\bibliographystyle{unsrt}
\bibliography{ref}
}

%%%%%%%%%%%%%%%%%%%%%%%%%%%%%%%%%%%%%%%%%%%%%%%%%%%%%%%%%%%%

\appendix

% \section{Technical Appendices and Supplementary Material}
% Technical appendices with additional results, figures, graphs and proofs may be submitted with the paper submission before the full submission deadline (see above), or as a separate PDF in the ZIP file below before the supplementary material deadline. There is no page limit for the technical appendices.

\section{Preliminary Concepts}\label{appendix:preliminary}
\subsection{Unified Task Definition and Notation}
We unify the formulation for robotic foundation models (including VLA and WAM) across discriminative and generative paradigms using consistent notation, focusing on the core mapping from multimodal inputs to task-compliant action sequences:
\begin{itemize}
    \item $\mathcal{I} = \{i_1, i_2, ..., i_T\}$: Sequence of visual observations (RGB images) at timesteps $1 \leq t \leq T$, where $i_t \in \mathbb{R}^{H \times W \times 3}$ in most cases.
    \item $L$: Natural language instruction (e.g., "stack the blue block on the red block"), tokenized to $L = [l_1, l_2, ..., l_N]$ via standard tokenizers (e.g., CLIP \cite{EVA_CLIP}, Bert \cite{devlin2019bert}).
    \item $\mathcal{A} = \{a_1, a_2, ..., a_T\}$: Robotic action sequence, with $a_t \in \mathbb{R}^D$ and $D$ denoting action dimensions (e.g., $D=7$ for \textsc{Libero} benchmarks).
    \item $\mathcal{A}^{gt}$: Ground-truth action sequence from expert demonstrations, serving as the target for model learning.
\end{itemize}

\subsection{Discriminative Robotic Foundation Models}
Discriminative paradigms \cite{kim2024openvla, kim2025fine, bu2025univla} cast robotic action learning as a deterministic regression task. Given visual observations $\mathcal{I}$ and language instructions $L$, a discriminative model $f_\theta$ (parameters $\theta$) directly predicts a deterministic action sequence:
\begin{equation}
    \label{eq:discriminative_vla}
    \mathcal{A}^{\text{pred}} = f_\theta(\mathcal{I}, L)
\end{equation}
Training minimizes the discrepancy between predicted and ground-truth actions via a regression loss over the dataset $\mathcal{D}$ (tuples of $\mathcal{I}, L, \mathcal{A}^\text{gt}$). Most existing works adopt an unweighted $\ell_1$ loss as the optimization objective:
\begin{equation}
    \label{eq:openvla_loss}
    \theta^* = \arg\min_\theta \mathbb{E}_{(\mathcal{I}, L, \mathcal{A}^\text{gt}) \sim \mathcal{D}}
    \left[ \frac{1}{T \cdot D} \sum_{t=1}^T \sum_{d=1}^D \left| a_{t,d}^{\text{pred}} - a_{t,d}^\text{gt} \right| \right]
\end{equation}

\subsection{Generative Robotic Foundation Models via Flow Matching}
The $\pi$-series models ($\pi_0$ \cite{black2024pi_0}, $\pi_{0.5}$ \cite{intelligence2025pi}) are representative generative frameworks based on flow matching. Instead of direct regression, these models learn a continuous vector field to transform Gaussian random noise $\epsilon \sim \mathcal{N}(\mathbf{0}, \mathbf{I})$ into task-aligned action sequences. Specifically, the learnable network $g_\phi$ predicts a time-dependent flow field conditioned on $\mathcal{I}$ and $L$, denoising random inputs toward $\mathcal{A}^\text{gt}$. The standard unweighted flow matching objective is:
\begin{align}
    \label{eq:flow_matching_pi0_pi05}
    \phi^* &= \arg\min_\phi \mathbb{E}_{\substack{(\mathcal{I}, L, \mathcal{A}^\text{gt}) \sim \mathcal{D} \\ \epsilon \sim \mathcal{N}(\mathbf{0}, \mathbf{I})}}
    \left[ \frac{1}{T \cdot D} \sum_{t=1}^T \sum_{d=1}^D
    \left\| u_t(\epsilon; \mathcal{I}, L) - \big(a_{t,d}^\text{gt} - \epsilon_d\big) \right\|_2^2 \right]
\end{align}
Here, $a_{t,d}^\text{gt}$ is the $d$-th dimension of $\mathcal{A}^\text{gt}$ at timestep $t$, with $T$ and $D$ denoting action sequence length and dimensionality, respectively. While $\pi_{0.5}$ introduces hierarchical reasoning for complex tasks, it retains the original flow matching objective. Related diffusion policy frameworks are discussed in Appendix~\ref{appendix:diff}.

\section{Action Attention Formulation of Generative VLA: Diffusion Policy}\label{appendix:diff}
\subsection{Generative VLA: Diffusion Model (Diffusion Policy)}
Diffusion Policy (DP) represents an independent line of generative work, distinct from $\pi_0$/$\pi_{0.5}$.
It models action generation as a iterative denoising process.
Given $K$ diffusion steps, the model $h_\psi$ predicts the noise $\epsilon_k^{\text{pred}}$ added to the action state at step $k$.
The standard unweighted $L_2$ diffusion optimization objective is:
\begin{align}
    \label{eq:diffusion_policy_loss}
    \psi^* &= \arg\min_\psi \mathbb{E}_{\substack{(\mathcal{I}, L, \mathcal{A}^{gt}) \sim \mathcal{D} \\ k \sim \text{Uniform}(1,K) \\ \epsilon_k \sim \mathcal{N}(0, I)}} \left[ \frac{1}{T \cdot D} \sum_{t=1}^T \sum_{d=1}^D \left\| \epsilon_k^{\text{pred}} - \epsilon_k \right\|_2^2 \right]
\end{align}
where $a_{t,d}^{(k)} = \alpha_k a_{t,d}^{gt} + \beta_k \epsilon_k$ is the noisy action, and $\alpha_k, \beta_k$ are pre-defined diffusion schedule coefficients.

\subsection{Revised Optimization Objective for Diffusion Policy (AttenA+Diff)}
For diffusion-based generative policies (Diffusion Policy), we apply the same velocity-based weighting to the denoising objective:
\begin{align}
    \label{eq:attenvla_diffusion_policy}
    \psi^* &= \arg\min_\psi \mathbb{E}_{\substack{(\mathcal{I}, L, \mathcal{A}^{gt}) \sim \mathcal{D} \\ k \sim \text{Uniform}(1,K) \\ \epsilon_k \sim \mathcal{N}(0, I)}} \left[ \frac{1}{T \cdot D} \sum_{t=1}^T \sum_{d=1}^D w_t \cdot \left\| \epsilon_k^{\text{pred}} - \epsilon_k \right\|_2^2 \right]
\end{align}
where $w_t$ emphasizes denoising accuracy for slow, critical timesteps during the diffusion process.

\section{Velocity-Based Action Attention Weighting Strategies}\label{appendix:weight}

We detail the four handcrafted weighting strategies used to implement velocity-field-based action attention in AttenA+.
These formulations serve as \emph{empirical, physics-inspired examples} to demonstrate the core idea of prioritizing slow, critical actions during training, rather than definitive or exclusive solutions.
All weighting rules assign higher importance to low-velocity timesteps, consistent with the intuition that precise manipulation phases require stricter optimization in our experiment datasets (\textsc{Libero}, RoboTwin 2.0).

For the $b$-th sample at timestep $t$, the velocity-aware weight $w_{b,t}$ is defined as follows:

\begin{enumerate}
    \item \textbf{Inverse strategy}
    \begin{equation}\label{eq:inverse}
        w_{b,t} = \frac{1}{v_{b,t}}
    \end{equation}
    This baseline scheme applies inverse weighting proportional to action speed, providing a mild but clear emphasis on slower movements.

    \item \textbf{Inverse squared strategy (amplified weight difference)}
    \begin{equation}\label{eq:inverse_squared}
        w_{b,t} = \frac{1}{v_{b,t}^2}
    \end{equation}
    By squaring the velocity term, this strategy strongly amplifies the contrast between slow and fast actions, making it the default choice in our main experiments.

    \item \textbf{Exponential decay strategy (fast attenuation)}
    \begin{equation}\label{eq:exp_decay}
        w_{b,t} = e^{-\alpha \cdot v_{b,t}}
    \end{equation}
    where $\alpha=5.0$ controls the decay rate. This method suppresses high-speed actions rapidly while maintaining soft weighting for slow segments.

    \item \textbf{Logarithmic strategy (smoothed weight)}
    \begin{equation}\label{eq:log}
        w_{b,t} = \frac{1}{\log(1 + v_{b,t})}
    \end{equation}
    The logarithmic transform yields gentle, stable weighting, reducing sensitivity to noise in velocity estimation.
\end{enumerate}

Notably, these four heuristic functions are \emph{example implementations} chosen for simplicity, interpretability, and empirical effectiveness.
They are not intended to limit the design space of action attention.
In future work, action weighting can be naturally extended to broader families of parametric functions, task-adaptive formulations, or \emph{fully learnable attention mechanisms} that infer importance end-to-end from data and physical constraints, rather than relying on fixed handcrafted rules.

\section{Visualization of Action Speed and Velocity-Guided Action Attention}
\label{app:visualization_attention}

In this section, we provide detailed analysis of the action speed patterns and velocity-guided attention weights, which are briefly summarized in the main text.

\begin{figure}
    \centering
    \includegraphics[width=1.0\linewidth]{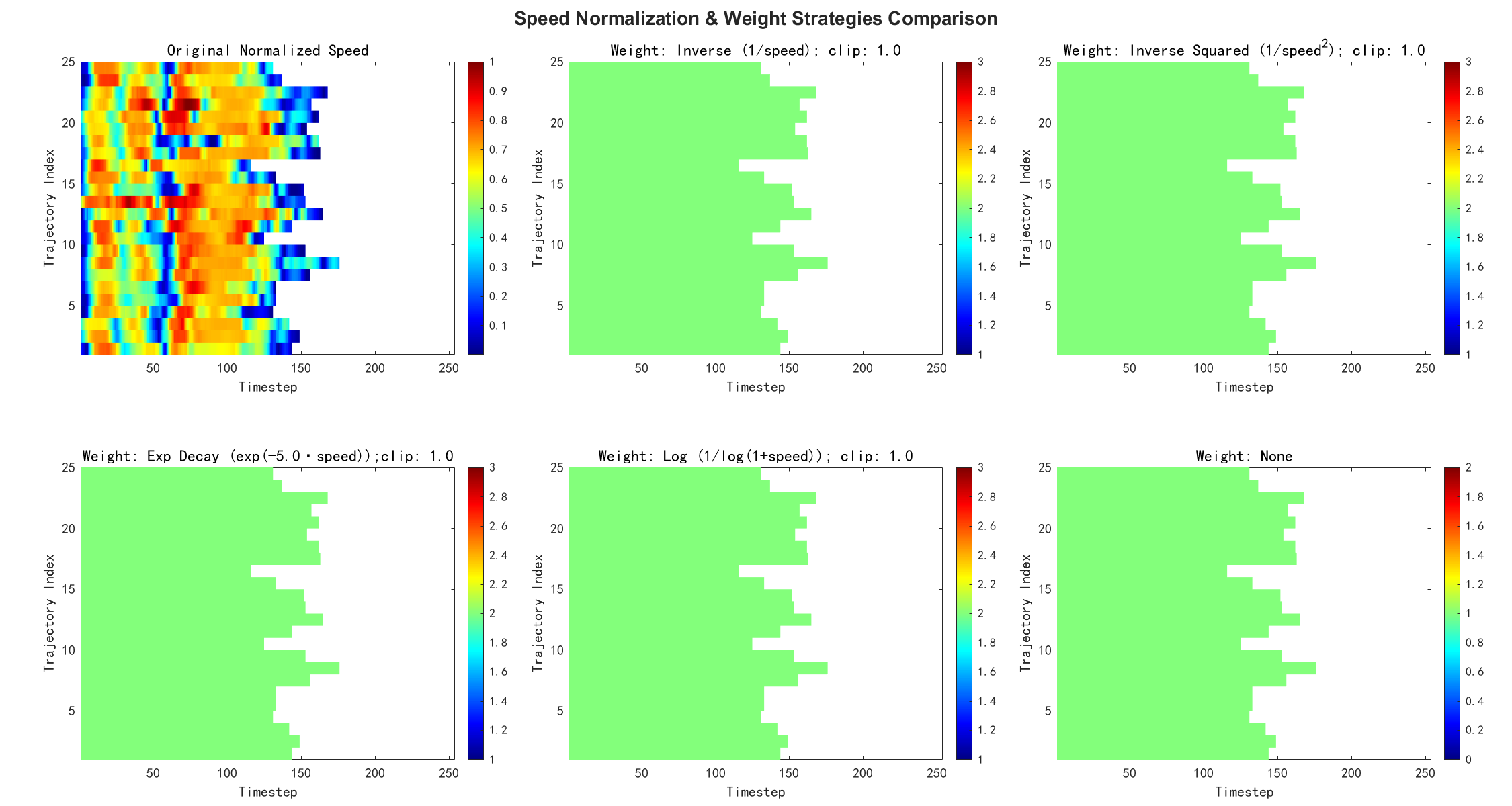}
    \caption{Visualization of Action Speed in \textsc{Libero-Object} Task with Different $clip_{max}=1.0$}
    \label{fig:object_clip01}
\end{figure}

\begin{figure}
    \centering
    \includegraphics[width=1.0\linewidth]{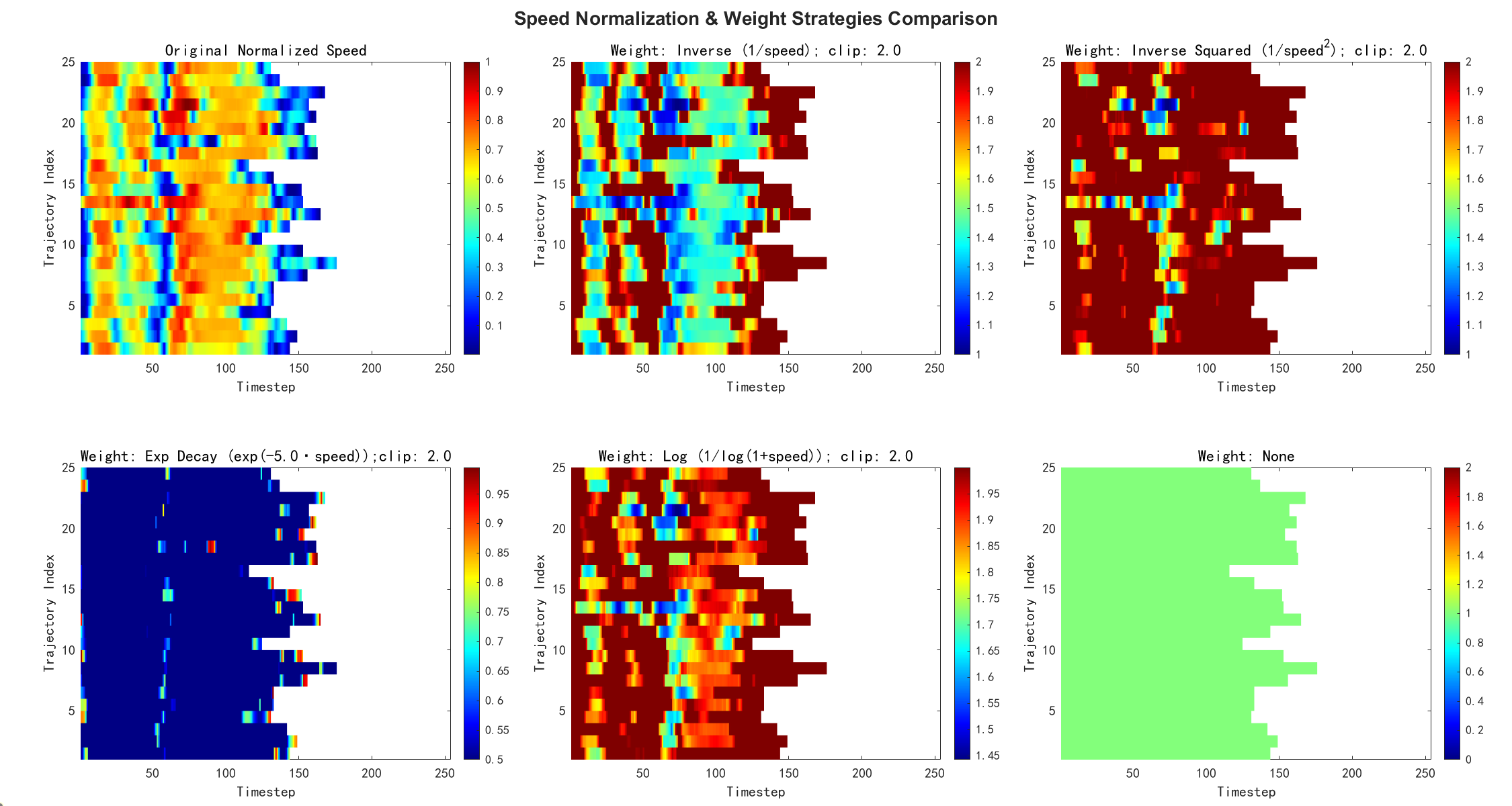}
    \caption{Visualization of Action Speed in \textsc{Libero-Object} Task with Different $clip_{max}=2.0$}
    \label{fig:object_clip02}
\end{figure}

\begin{figure}
    \centering
    \includegraphics[width=1.0\linewidth]{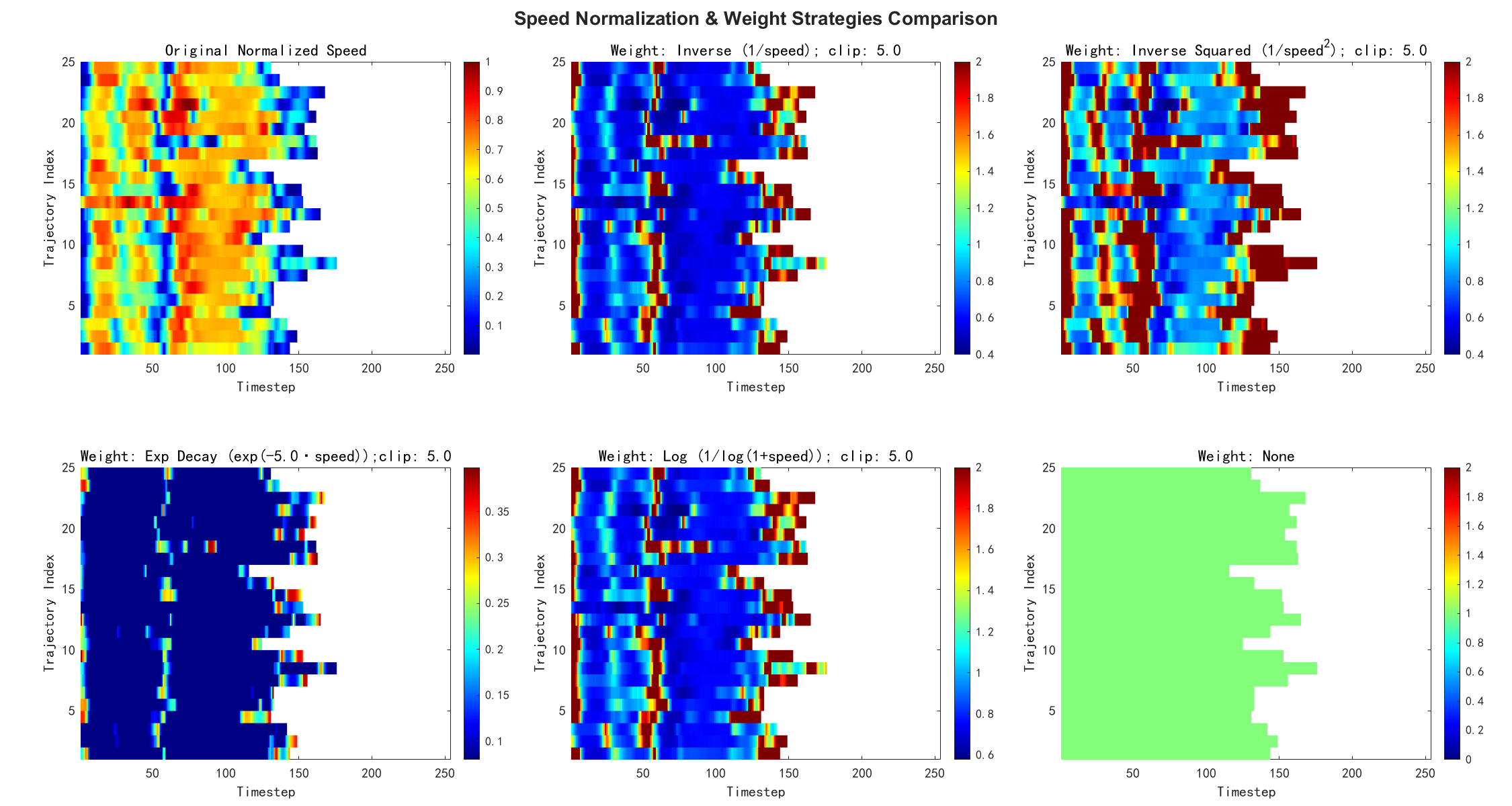}
    \caption{Visualization of Action Speed in \textsc{Libero-Object} Task with Different $clip_{max}=5.0$}
    \label{fig:object_clip05}
\end{figure}

\begin{figure}
    \centering
    \includegraphics[width=1.0\linewidth]{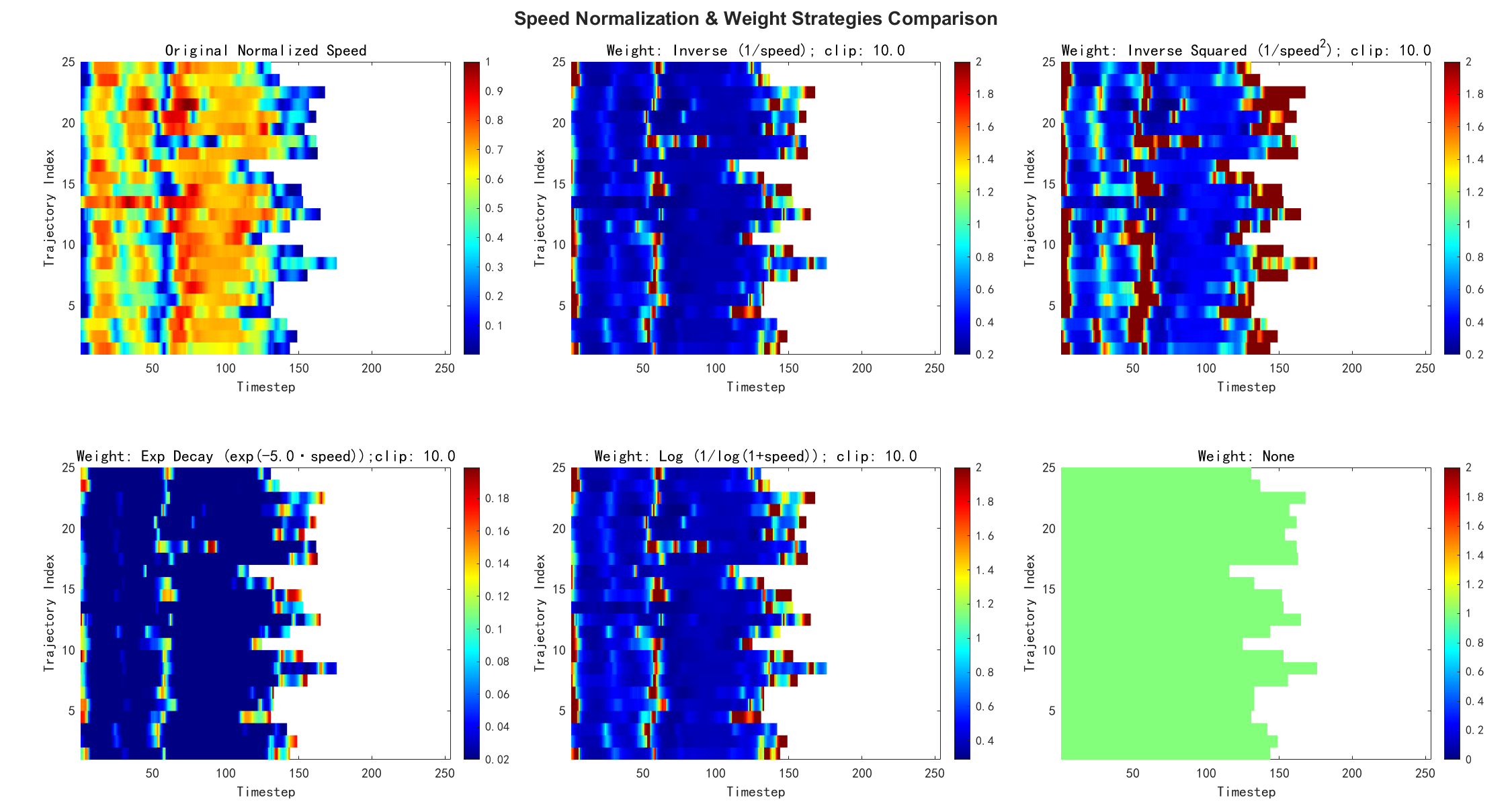}
    \caption{Visualization of Action Speed in \textsc{Libero-Object} Task with Different $clip_{max}=10.0$}
    \label{fig:object_clip10}
\end{figure}

Figures \ref{fig:object_clip01}–\ref{fig:object_clip10} present comprehensive visualizations of raw action velocity profiles and the resulting velocity-field-based attention weights under four distinct clipping thresholds \(clip_{\text{max}} \in \{1.0, 2.0, 5.0, 10.0\}\) on the \textsc{Libero-object} manipulation benchmark.
Each figure follows a consistent layout: Subplot 1 illustrates the temporal distribution of raw action speed magnitudes across multiple expert demonstration trajectories, revealing inherent speed variations within task execution.
Subplots 2 through 5 display the attention weight distributions generated by the four velocity transformation rules (Equations \ref{eq:inverse}--\ref{eq:log}): inverse weighting, inverse squared weighting, exponential decay weighting, and logarithmic weighting, respectively.
Subplot 6 serves as the baseline, showing the original action trajectory under uniform, unweighted treatment without action attention.

From the raw velocity visualizations, distinct slow–fast motion patterns emerge consistently across all task trajectories.
Slow-motion segments consistently align with task-critical phases, including robot initialization, precise object approaching, fine manipulation, grasping, targeted placement, and task completion.
These phases demand high positional accuracy and are highly sensitive to execution errors, making them decisive for overall task success.
Conversely, fast-motion segments correspond to robust, error-tolerant transitional movements, such as free-space arm traversal, coarse positioning toward target objects, and post-grasp repositioning, where minor deviations rarely lead to task failure.

The effect of the clipping threshold \(clip_{\text{max}}\) is clearly demonstrated across Figures \ref{fig:object_clip01}–\ref{fig:object_clip10}.
At \(clip_{\text{max}} = 1.0\), all velocity-adaptive weighting schemes collapse to uniform values, reverting AttenA+ to a standard VLA model with equal emphasis on all timesteps.
As \(clip_{\text{max}}\) increases sequentially from 1.0 to 2.0, 5.0, and 10.0, the discriminative power of action attention is progressively strengthened: slow critical actions receive increasingly prominent weights, while fast transitional actions are assigned relatively lower weights, widening the gap in learning priority.

Moreover, the four velocity mapping functions exhibit distinctive attention characteristics.
As clearly visible in the \(clip_{\text{max}} = 2.0\) visualization, exponential decay weighting (Equation \ref{eq:exp_decay}) produces highly localized emphasis: it strongly amplifies a small set of extremely slow actions while broadly suppressing fast actions across a wide range.
In contrast, inverse (Equation \ref{eq:inverse}), inverse squared (Equation \ref{eq:inverse_squared}), and logarithmic (Equation \ref{eq:log}) schemes maintain widespread emphasis on slow actions and exert mild, localized suppression on fast actions.
Within this group, the intensity of low-speed amplification follows a clear hierarchy: inverse squared (Equation \ref{eq:inverse_squared}) yields the strongest enhancement, followed by logarithmic weighting (Equation \ref{eq:log}), and then inverse weighting (Equation \ref{eq:inverse}).
This consistent trend is observable across all clipping thresholds in Figures \ref{fig:object_clip01}–\ref{fig:object_clip10}, validating the design principles of velocity-field-based action attention and supporting the selection of inverse squared weighting as the default configuration in the main experiments.

\section{Details about Model Training}\label{appendix:training}

This section presents comprehensive training and implementation details for AttenA+OFT (evaluated on \textsc{LIBERO}) and AttenA+WAM (evaluated on RoboTwin), covering architectural modifications, optimization configurations, fine-tuning pipelines, checkpoint scheduling, and best-model selection criteria.
All experiments in this work use the weight clipping settings $clip_{\text{max}}=2.0$ and $clip_{\text{max}}=5.0$, which are applied consistently across both model variants.

\subsection{AttenA+OFT}\label{appendix:oft}
We build AttenA+OFT as a direct adaptation of the OpenVLA-OFT framework, with our core velocity-field action attention integrated as a weighted module without altering the backbone architecture.
Following the standard OpenVLA-OFT fine-tuning protocol, we train separate models for each of the four \textsc{LIBERO} task categories (Spatial, Object, Goal, Long) to ensure fair comparison with prior work.
All models are trained for a total of $200,\!000$ steps, with checkpoints saved every $5,\!000$ steps. The taining time of each model using single H800 GPU is about 35 hours.
During training, we strictly retain the original optimizer configuration, learning rate schedule, batch size, and data preprocessing used in OpenVLA-OFT to isolate the improvement brought by action attention.
After training, we evaluate over saved checkpoints on the corresponding \textsc{LIBERO} test split and select the checkpoint with the highest success rate as the final model for reporting results.

\subsection{AttenA+WAM}\label{appendix:wam}
We implement AttenA+WAM on top of the Fast-WAM architecture, again integrating our velocity-guided action attention as a plug-and-play weighting module.
Since the official Fast-WAM repository does not release end-to-end fine-tuning code, we adopt a practical and fair adaptation protocol: we freeze all vision encoders and the pre-trained WAM backbone, and only fine-tune the final action head using our proposed action attention mechanism.
This design ensures we only introduce our method while preserving the pre-trained knowledge of the original model.
We fine-tune on the RoboTwin dataset for 1 epoch, saving checkpoints every $2,\!000$ steps. The taining time of the model using two H800 GPUs is about 4 days.
Consistent with AttenA+OFT, we evaluate over intermediate checkpoints and select the best-performing one based on validation success rate for final experimental comparisons.

\section{Detailed Evaluation Results on RoboTwin 2.0 and Real Franka Robot Experiments}
\label{appendix:robotwin}

\subsection{Detailed Results on RoboTwin 2.0}
\begin{table*}[t]
\centering
\caption{We present quantitative results on the RoboTwin 2.0 simulation benchmark, covering 50 bimanual manipulation tasks with two difficulty levels. RoboTwin 2.0 serves as a rigorous dual-arm manipulation testbed that demands precise bilateral coordination. The easy setting adopts fixed initial scene arrangements, whereas the hard setting introduces randomized object placements and scene configurations for higher generalization challenges.}
\resizebox{\textwidth}{!}{%
\begin{tabular}{lcccccccccccccc}
\toprule
\multirow{2}{*}{Model} 
& \multicolumn{2}{c}{AttenA+WAM(Ours)} 
& \multicolumn{2}{c}{Fast-WAM} 
& \multicolumn{2}{c}{LingBot} 
& \multicolumn{2}{c}{Pi\_05} 
& \multicolumn{2}{c}{Pi\_0} 
& \multicolumn{2}{c}{X-VLA} 
& \multicolumn{2}{c}{Motus} \\
\cmidrule(lr){2-3} \cmidrule(lr){4-5} \cmidrule(lr){6-7} 
\cmidrule(lr){8-9} \cmidrule(lr){10-11} \cmidrule(lr){12-13} \cmidrule(lr){14-15}
Task Type 
& clean & random & clean & random & clean & random 
& clean & random & clean & random & clean & random & clean & random \\
\midrule
Adjust Bottle & 100 & 100 & 100 & 100 & 90 & 94 & 100 & 99 & 99 & 95 & 100 & 99 & 89 & 93 \\
Beat Block Hammer & 99 & 93 & 99 & 97 & 96 & 98 & 96 & 93 & 79 & 84 & 92 & 88 & 95 & 88 \\
Blocks Ranking RGB & 100 & 100 & 100 & 100 & 99 & 98 & 92 & 85 & 80 & 63 & 83 & 83 & 99 & 97 \\
Blocks Ranking Size & 93 & 94 & 94 & 98 & 94 & 96 & 49 & 26 & 14 & 5 & 67 & 74 & 75 & 63 \\
Click Alarmclock & 100 & 100 & 100 & 100 & 99 & 100 & 98 & 89 & 77 & 68 & 99 & 99 & 100 & 100 \\
Click Bell & 100 & 100 & 100 & 100 & 100 & 100 & 99 & 66 & 71 & 48 & 100 & 100 & 100 & 100 \\
Dump Bin Big Binbin & 97 & 95 & 97 & 96 & 89 & 96 & 92 & 97 & 88 & 83 & 79 & 77 & 95 & 91 \\
Grab Roller & 100 & 100 & 100 & 100 & 100 & 100 & 100 & 100 & 98 & 94 & 100 & 100 & 100 & 100 \\
Handover Block & 95 & 90 & 95 & 81 & 99 & 78 & 66 & 57 & 47 & 31 & 73 & 37 & 86 & 73 \\
Handover Mic & 100 & 91 & 99 & 100 & 94 & 96 & 98 & 97 & 97 & 97 & 0 & 0 & 78 & 63 \\
Hanging Mug & 67 & 62 & 58 & 62 & 40 & 28 & 18 & 17 & 14 & 11 & 23 & 27 & 38 & 38 \\
Lift Pot & 100 & 100 & 100 & 100 & 100 & 99 & 96 & 85 & 80 & 72 & 99 & 100 & 96 & 99 \\
Move Can Pot & 89 & 91 & 90 & 88 & 94 & 97 & 51 & 55 & 68 & 48 & 89 & 86 & 34 & 74 \\
Move Pillowbottle Pad & 98 & 100 & 100 & 99 & 99 & 99 & 84 & 61 & 67 & 46 & 73 & 71 & 93 & 96 \\
Move Playingcard Away & 100 & 100 & 100 & 100 & 100 & 99 & 96 & 84 & 74 & 65 & 93 & 98 & 100 & 96 \\
Move Stapler Pad & 74 & 70 & 77 & 64 & 91 & 79 & 56 & 42 & 41 & 24 & 78 & 73 & 83 & 85 \\
Open Laptop & 99 & 100 & 98 & 100 & 92 & 94 & 90 & 96 & 71 & 81 & 93 & 100 & 95 & 91 \\
Open Microwave & 71 & 49 & 62 & 45 & 82 & 86 & 34 & 77 & 4 & 32 & 79 & 71 & 95 & 91 \\
Pick Diverse Bottles & 91 & 85 & 80 & 85 & 89 & 82 & 81 & 71 & 69 & 31 & 58 & 36 & 90 & 91 \\
Pick Dual Bottles & 100 & 96 & 100 & 96 & 100 & 99 & 93 & 63 & 59 & 37 & 47 & 36 & 96 & 90 \\
Place A2B Left & 97 & 96 & 95 & 93 & 97 & 93 & 87 & 82 & 43 & 47 & 48 & 49 & 82 & 79 \\
Place A2B Right & 95 & 98 & 93 & 99 & 97 & 95 & 87 & 84 & 39 & 34 & 36 & 36 & 90 & 87 \\
Place Bread Basket & 93 & 93 & 91 & 93 & 97 & 95 & 77 & 64 & 62 & 46 & 81 & 71 & 91 & 94 \\
Place Bread Skillet & 88 & 91 & 90 & 93 & 95 & 90 & 85 & 66 & 66 & 49 & 77 & 67 & 86 & 83 \\
Place Burger Fries & 96 & 96 & 96 & 99 & 97 & 95 & 94 & 87 & 81 & 76 & 94 & 94 & 98 & 98 \\
Place Can Basket & 65 & 64 & 71 & 69 & 81 & 84 & 62 & 62 & 55 & 46 & 49 & 52 & 81 & 76 \\
Place Cans Plasticbox & 99 & 100 & 99 & 96 & 100 & 99 & 94 & 84 & 63 & 45 & 97 & 98 & 98 & 94 \\
Place Container Plate & 98 & 99 & 96 & 100 & 99 & 97 & 99 & 95 & 97 & 92 & 97 & 95 & 98 & 99 \\
Place Dual Shoes & 86 & 89 & 94 & 88 & 94 & 89 & 75 & 75 & 59 & 51 & 79 & 88 & 93 & 87 \\
Place Empty Cup & 99 & 100 & 100 & 100 & 100 & 100 & 100 & 99 & 91 & 85 & 100 & 98 & 99 & 98 \\
Place Fan & 97 & 91 & 96 & 96 & 99 & 93 & 87 & 85 & 66 & 71 & 80 & 75 & 91 & 87 \\
Place Mouse Pad & 88 & 88 & 83 & 89 & 93 & 96 & 60 & 39 & 20 & 20 & 70 & 70 & 66 & 68 \\
Place Object Basket & 91 & 85 & 89 & 88 & 91 & 88 & 80 & 76 & 67 & 70 & 44 & 39 & 81 & 87 \\
Place Object Scale & 92 & 94 & 90 & 97 & 96 & 95 & 86 & 80 & 57 & 52 & 52 & 74 & 88 & 85 \\
Place Object Stand & 90 & 91 & 90 & 94 & 99 & 96 & 91 & 85 & 82 & 68 & 86 & 88 & 98 & 97 \\
Place Phone Stand & 99 & 99 & 97 & 99 & 97 & 97 & 81 & 81 & 49 & 53 & 88 & 87 & 87 & 86 \\
Place Shoe & 96 & 99 & 96 & 99 & 98 & 98 & 92 & 93 & 76 & 76 & 96 & 95 & 99 & 97 \\
Press Stapler & 92 & 94 & 90 & 97 & 85 & 82 & 87 & 83 & 44 & 37 & 92 & 98 & 93 & 98 \\
Put Bottles Dustbin & 91 & 91 & 95 & 90 & 87 & 91 & 84 & 79 & 65 & 56 & 74 & 77 & 81 & 79 \\
Put Object Cabinet & 85 & 85 & 94 & 89 & 85 & 87 & 80 & 79 & 73 & 60 & 46 & 48 & 88 & 71 \\
Rotate QRcode & 92 & 91 & 93 & 89 & 96 & 91 & 89 & 87 & 74 & 70 & 34 & 33 & 89 & 73 \\
Scan Object & 93 & 89 & 89 & 92 & 96 & 91 & 72 & 65 & 55 & 42 & 14 & 36 & 67 & 66 \\
Shake Bottle Horizontally & 100 & 100 & 100 & 100 & 100 & 99 & 99 & 99 & 98 & 92 & 100 & 100 & 100 & 98 \\
Shake Bottle & 100 & 100 & 100 & 100 & 100 & 97 & 99 & 97 & 94 & 91 & 99 & 100 & 100 & 97 \\
Stack Blocks Three & 97 & 96 & 95 & 97 & 99 & 98 & 91 & 76 & 72 & 52 & 6 & 10 & 91 & 95 \\
Stack Blocks Two & 100 & 100 & 100 & 100 & 100 & 98 & 97 & 100 & 93 & 79 & 92 & 87 & 100 & 98 \\
Stack Bowls Three & 95 & 95 & 80 & 81 & 86 & 83 & 77 & 71 & 77 & 75 & 76 & 86 & 79 & 87 \\
Stack Bowls Two & 95 & 95 & 92 & 98 & 94 & 98 & 95 & 96 & 94 & 95 & 96 & 93 & 98 & 98 \\
Stamp Seal & 91 & 89 & 90 & 94 & 96 & 97 & 79 & 55 & 46 & 33 & 76 & 82 & 93 & 92 \\
Turn Switch & 80 & 79 & 61 & 59 & 44 & 45 & 62 & 54 & 41 & 42 & 40 & 61 & 84 & 78 \\
\midrule
Average & 93.06 & 91.86 & 91.88 & 91.78 & 92.9 & 91.5 & 82.74 & 76.76 & 65.92 & 58.4 & 72.88 & 72.84 & 88.52 & 87.02 \\
\bottomrule
\end{tabular}
}
\label{tab:robo_twin_sim}
\end{table*}

% \subsection{Detailed Evaluation Results on Franka Robot}

\begin{table}[t]
\centering
\caption{Detailed Evaluation Results on Real Franka Robot Experiments. We report the success rates (SR) across four manipulation tasks under the Baseline and our Attention-based method.}
\vspace{0.5em}
\resizebox{0.95\linewidth}{!}{%
\begin{tabular}{@{}l cccccccc@{}}
\toprule
Experiment & \multicolumn{4}{c}{Baseline} & \multicolumn{4}{c}{AttenA+} \\
\cmidrule(lr){2-5} \cmidrule(lr){6-9}
& Close Draw & Put Cube & Multi-object & Long & Close Draw & Put Cube & Multi-object & Long \\
\midrule
1  & 1 & 1 & 1 & 1 & 1 & 1 & 1 & 1 \\
2  & 1 & 1 & 1 & 0 & 1 & 1 & 1 & 1 \\
3  & 1 & 1 & 1 & 1 & 1 & 1 & 1 & 1 \\
4  & 1 & 1 & 1 & 1 & 1 & 1 & 1 & 1 \\
5  & 1 & 1 & 1 & 1 & 1 & 1 & 1 & 1 \\
6  & 1 & 1 & 0 & 1 & 1 & 1 & 1 & 1 \\
7  & 1 & 1 & 1 & 0 & 1 & 1 & 1 & 1 \\
8  & 1 & 1 & 1 & 1 & 1 & 1 & 1 & 1 \\
9  & 1 & 1 & 1 & 1 & 1 & 1 & 1 & 0 \\
10 & 1 & 1 & 1 & 1 & 1 & 1 & 1 & 1 \\
11 & 1 & 1 & 1 & 1 & 1 & 1 & 1 & 1 \\
12 & 1 & 0 & 1 & 0 & 1 & 1 & 1 & 1 \\
13 & 1 & 1 & 1 & 0 & 1 & 1 & 1 & 1 \\
14 & 1 & 1 & 0 & 1 & 1 & 1 & 1 & 1 \\
15 & 1 & 1 & 1 & 1 & 1 & 1 & 1 & 1 \\
16 & 1 & 1 & 0 & 1 & 1 & 1 & 1 & 1 \\
17 & 1 & 1 & 1 & 1 & 1 & 1 & 1 & 1 \\
18 & 1 & 1 & 1 & 0 & 1 & 1 & 1 & 1 \\
19 & 1 & 1 & 1 & 1 & 1 & 1 & 1 & 1 \\
20 & 1 & 1 & 1 & 1 & 1 & 1 & 1 & 1 \\
21 & 1 & 1 & 1 & 1 & 1 & 1 & 1 & 1 \\
22 & 1 & 1 & 1 & 1 & 1 & 1 & 1 & 0 \\
23 & 1 & 1 & 1 & 1 & 1 & 1 & 1 & 1 \\
24 & 1 & 1 & 1 & 1 & 1 & 1 & 1 & 1 \\
25 & 1 & 1 & 1 & 1 & 1 & 1 & 1 & 1 \\
26 & 1 & 1 & 1 & 1 & 1 & 1 & 1 & 1 \\
27 & 1 & 1 & 1 & 1 & 1 & 1 & 1 & 1 \\
28 & 1 & 1 & 1 & 1 & 1 & 1 & 1 & 1 \\
29 & 1 & 1 & 1 & 1 & 1 & 1 & 0 & 1 \\
30 & 1 & 1 & 1 & 1 & 1 & 1 & 1 & 0 \\
31 & 1 & 1 & 1 & 1 & 1 & 1 & 1 & 1 \\
32 & 1 & 1 & 1 & 1 & 1 & 1 & 1 & 1 \\
33 & 1 & 1 & 1 & 1 & 1 & 1 & 1 & 1 \\
34 & 1 & 1 & 0 & 1 & 1 & 1 & 1 & 1 \\
35 & 1 & 1 & 1 & 1 & 1 & 1 & 1 & 1 \\
36 & 1 & 1 & 1 & 1 & 1 & 1 & 1 & 1 \\
37 & 1 & 1 & 1 & 1 & 1 & 1 & 1 & 1 \\
38 & 1 & 0 & 1 & 1 & 1 & 1 & 1 & 1 \\
39 & 1 & 1 & 1 & 0 & 1 & 1 & 1 & 1 \\
40 & 1 & 1 & 1 & 1 & 1 & 1 & 1 & 1 \\
41 & 1 & 1 & 1 & 1 & 1 & 1 & 1 & 0 \\
42 & 1 & 1 & 1 & 1 & 1 & 1 & 1 & 1 \\
43 & 1 & 1 & 0 & 0 & 1 & 1 & 1 & 1 \\
44 & 1 & 1 & 1 & 1 & 1 & 1 & 1 & 0 \\
45 & 1 & 1 & 1 & 1 & 1 & 1 & 1 & 1 \\
46 & 1 & 1 & 1 & 1 & 1 & 1 & 1 & 1 \\
47 & 1 & 1 & 1 & 0 & 1 & 1 & 1 & 1 \\
48 & 1 & 1 & 1 & 1 & 1 & 1 & 1 & 1 \\
49 & 1 & 1 & 1 & 1 & 1 & 1 & 1 & 1 \\
50 & 1 & 1 & 1 & 1 & 1 & 1 & 1 & 1 \\
\midrule
\textbf{SR (\%)} & \textbf{100} & \textbf{96} & \textbf{90} & \textbf{84} & \textbf{100} & \textbf{100} & \textbf{98} & \textbf{90} \\
\bottomrule
\end{tabular}%
}
\label{tab:franka_real_detailed}
\end{table}

% \begin{figure}
%     \centering
%     \includegraphics[width=1.0\linewidth]{figures/appendix_sim_libero_experiments.png}
%     \caption{Third views of 4 \textsc{Libero} Tasks. We add `critical' label to show the frame of slow actions which is critical for the success of the task. }
%     \label{fig:appendix_sim_libero_experiments}
% \end{figure}

% \begin{figure}
%     \centering
%     \includegraphics[width=1.0\linewidth]{figures/appendix_sim_robotwin_experiments.png}
%     \caption{Third views of example RoboTwin Clean/Random Tasks. We add `critical' label to show the frame of slow actions which is critical for the success of the task. }
%     \label{fig:appendix_sim_robotwin_experiments}
% \end{figure}

% \begin{figure}
%     \centering
%     \includegraphics[width=1.0\linewidth]{figures/appendix_real_experiments.png}
%     \caption{Third views of 4 Real Tasks. We add `critical' label to show the frame of slow actions which is critical for the success of the task. }
%     \label{fig:appendix_real_experiments}
% \end{figure}
\begin{figure}
    \centering
    \includegraphics[width=1.0\linewidth]{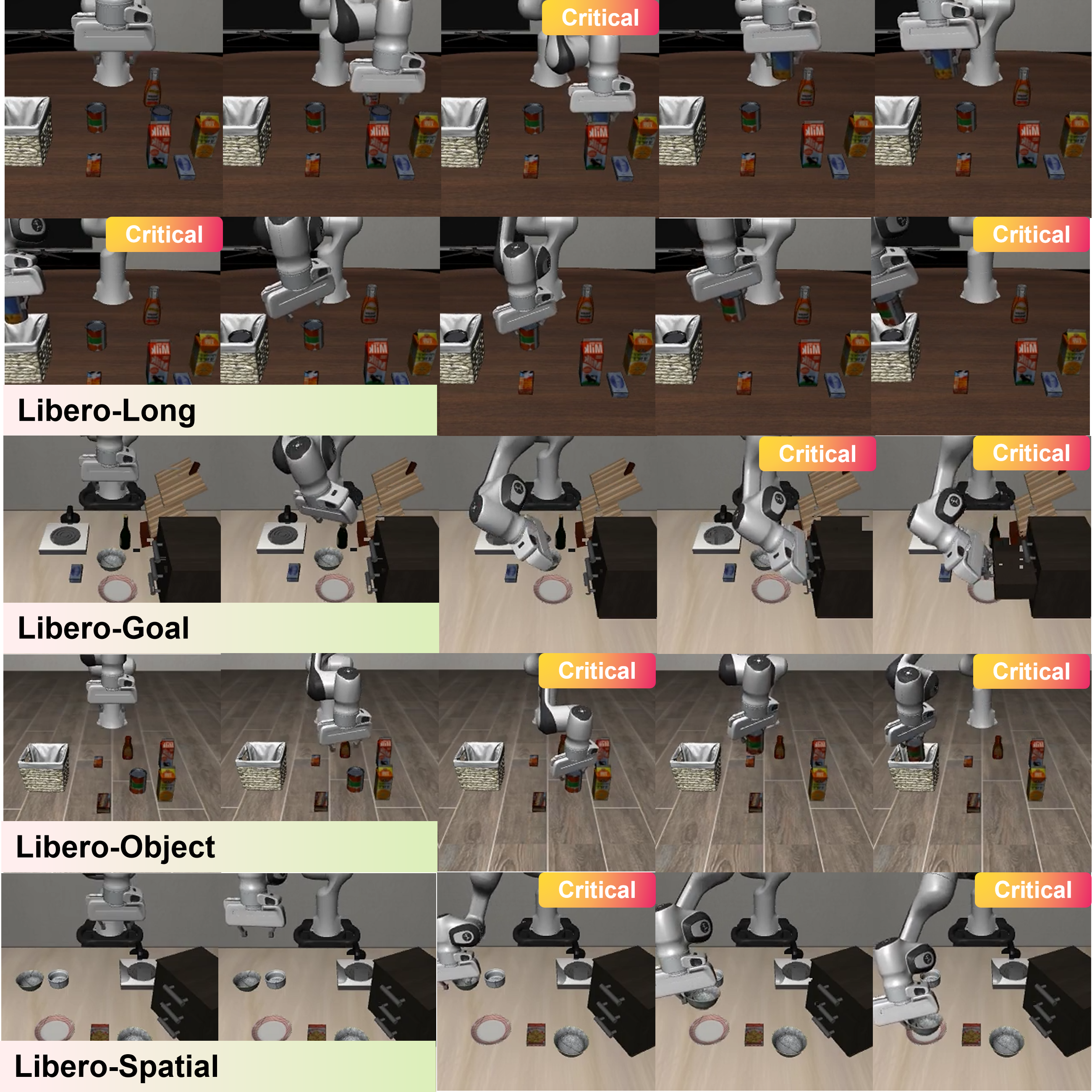}
    \caption{Third-person views of example \textsc{Libero} manipulation tasks. Frames labeled `critical' highlight slow, high-precision actions (e.g., grasping, alignment) where AttenA+ applies increased attention weights to improve task success.}
    \label{fig:appendix_sim_libero_experiments}
\end{figure}

\begin{figure}
    \centering
    \includegraphics[width=1.0\linewidth]{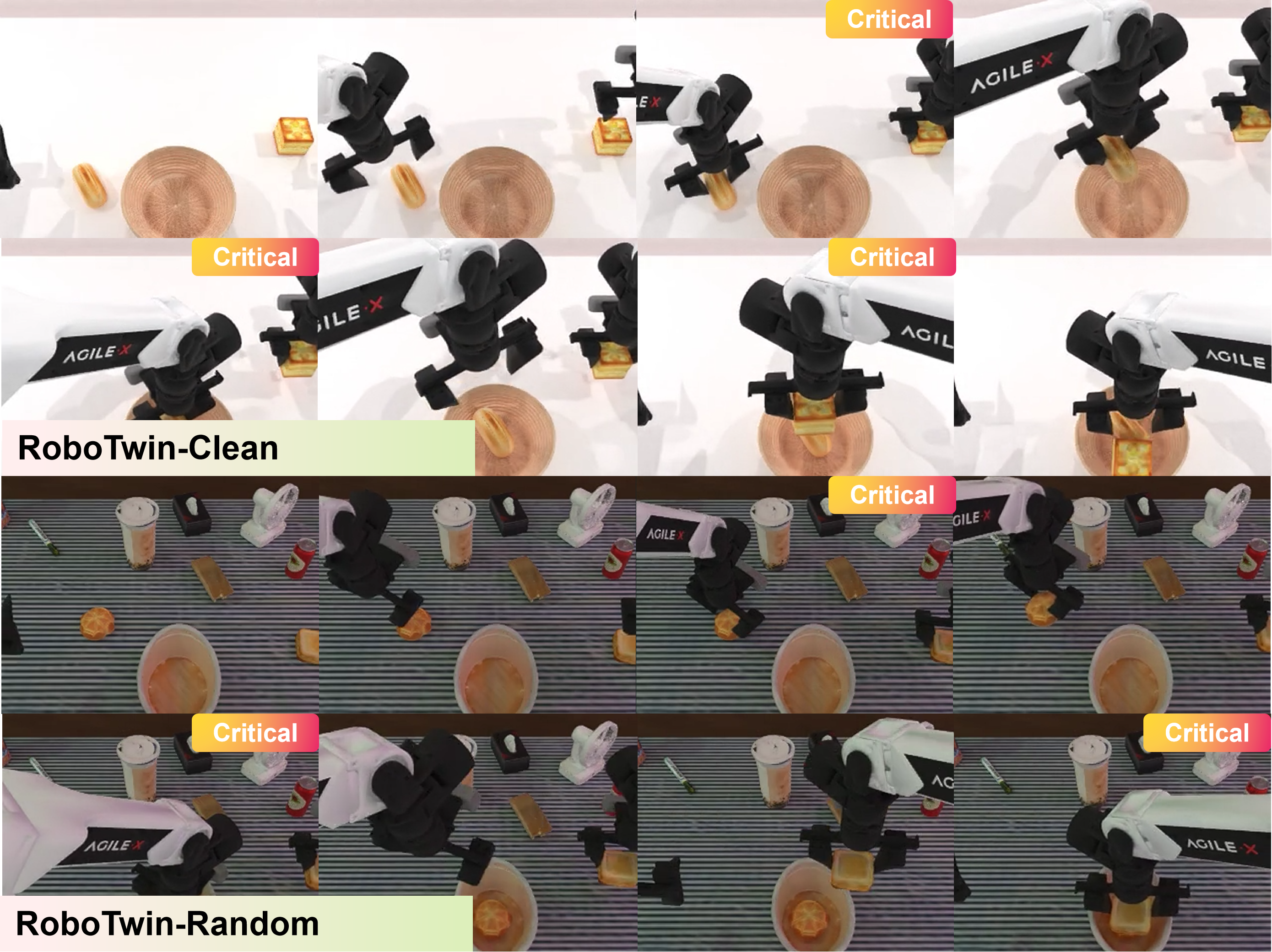}
    \caption{Third-person views of example RoboTwin tasks in both clean and randomized environments. The `critical' labels mark slow, precision-sensitive steps, where AttenA+ prioritizes learning to boost performance across diverse conditions.}
    \label{fig:appendix_sim_robotwin_experiments}
\end{figure}

\begin{figure}
    \centering
    \includegraphics[width=1.0\linewidth]{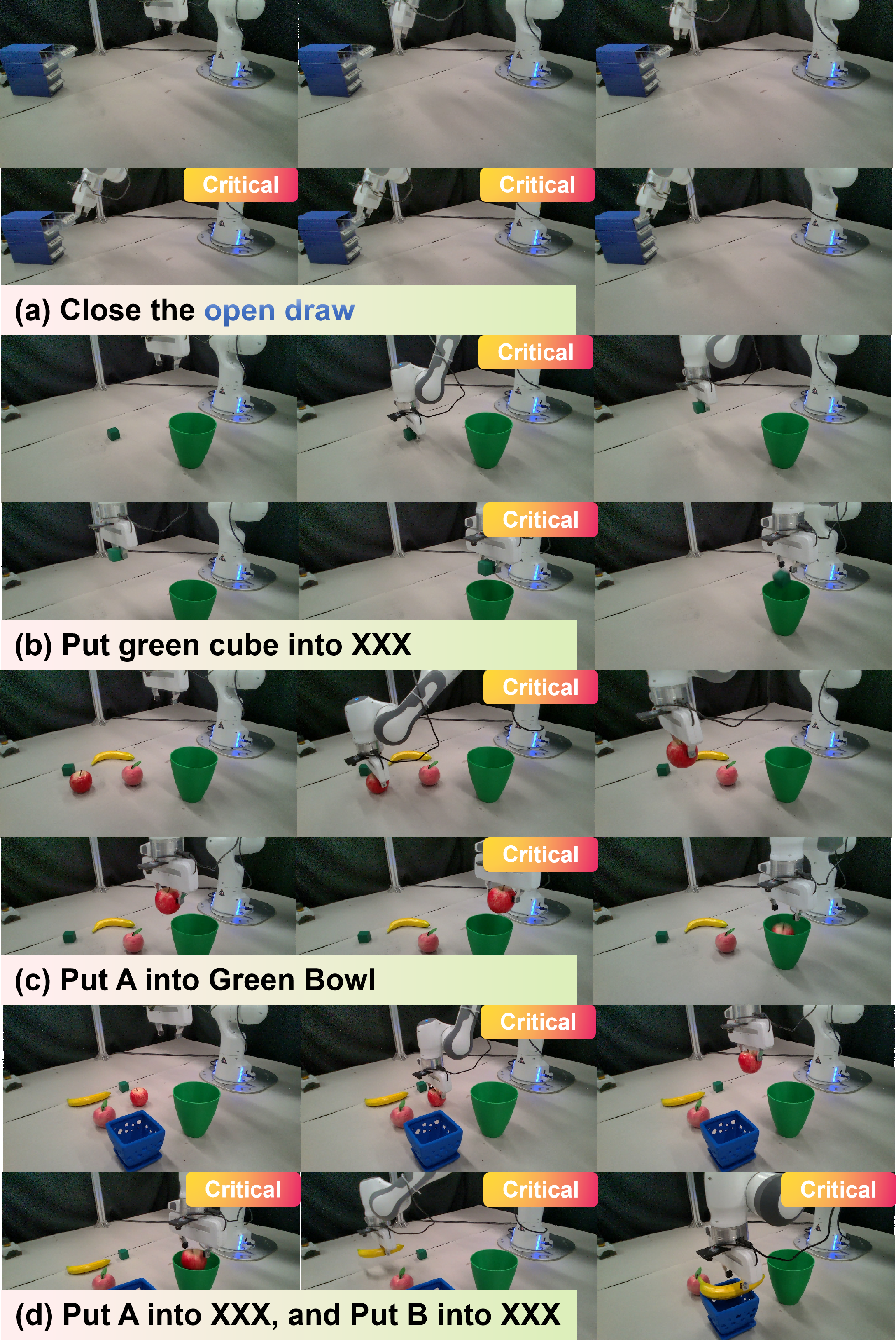}
    \caption{Third-person views of four representative real-world Franka tasks. The `critical' labels identify slow, high-precision manipulation steps, demonstrating how AttenA+ prioritizes these phases to improve real-robot success rates.}
    \label{fig:appendix_real_experiments}
\end{figure}

\end{document}